\pdfoutput=1
\documentclass[11pt]{article}
\usepackage{ACL2023}
\usepackage{CJKutf8}
\usepackage{times}
\usepackage{color}
\usepackage{array}
\usepackage{enumitem}
\usepackage{caption}
\usepackage{subfig}
\usepackage{algorithm}
\usepackage{algorithmic}
\usepackage{amsmath}
\usepackage{amssymb}
\usepackage{utfsym}
\usepackage{pifont}
\usepackage{latexsym}
\usepackage{graphicx}
\usepackage{multirow}
\usepackage{booktabs}
\usepackage{arydshln}
\usepackage{bm}
\usepackage{tikz}
\usepackage{pgfplots}
\usepackage{ragged2e}
\usepackage{makecell}
\pgfplotsset{compat=1.8}
\usepackage[T1]{fontenc}
\usepackage[utf8]{inputenc}
\usepackage{microtype}
\usepackage{times}
\usepackage{latexsym}
\usepackage[T1]{fontenc}
\usepackage[utf8]{inputenc}
\usepackage{microtype}
\usepackage{inconsolata}
\usepackage{paralist, tabularx}
\usepackage{xspace}
\usepackage[utf8]{inputenc}
\usepackage{latexsym}
\usepackage{inconsolata}

\definecolor{myblue}{RGB}{0,191,255}
\definecolor{mygreen}{RGB}{0,204,102}
\definecolor{winered}{RGB}{232,84,84}
\definecolor{mypurple}{RGB}{153,153,255}
\newcommand{\myroman}[1]{{\uppercase\expandafter{\romannumeral#1}}}
\newcommand{\xdashleftrightarrow}[2][]{\ext@arrow 3359\leftrightarrowfill@@{#1}{#2}}
\definecolor{ired}{RGB}{229,72,72}
\definecolor{igreen}{RGB}{80,219,144}

\newcommand{\mlogo}{\raisebox{-1pt}{\includegraphics[width=2em]{figure/logo3.pdf}}\xspace}

\title{\textit{DiaASQ}\mlogo: A Benchmark of Conversational\\ Aspect-based Sentiment Quadruple Analysis}

\author{
Bobo Li\textsuperscript{\rm 1}, \,
Hao Fei\textsuperscript{\rm 2},  \,
Fei Li\textsuperscript{\rm 1}, \,
Yuhan Wu\textsuperscript{\rm 1}, \,
Jinsong Zhang\textsuperscript{\rm 1}, \,
Shengqiong Wu\textsuperscript{\rm 2}, \\
\textbf{
Jingye Li\textsuperscript{\rm 1},  \,
Yijiang Liu\textsuperscript{\rm 1},  \,
Lizi Liao\textsuperscript{\rm 3},  \,
Tat-Seng Chua\textsuperscript{\rm 2} and Donghong Ji\textsuperscript{\rm 1}\thanks{\ \ Corresponding author.}
}\\
\textsuperscript{\rm 1} Key Laboratory of Aerospace Information Security and Trusted Computing, Ministry  \\of Education,
School of Cyber Science and Engineering, Wuhan University \\
\textsuperscript{\rm 2} Sea-NExT Joint Lab, National University of Singapore \, 
\textsuperscript{\rm 3} Singapore Management University 
\\
\texttt{\{boboli,lifei\_csnlp,yuhanwu,jinsongzhang,theodorelee,cslyj,dhji\}@whu.edu.cn}
\\
\texttt{\{haofei37,dcscts\}@nus.edu.sg}\quad
\texttt{swu@u.nus.edu}\quad
\texttt{lzliao@smu.edu.sg}
}

\begin{document}

\maketitle
\begin{abstract}
The rapid development of aspect-based sentiment analysis (ABSA) within recent decades shows great potential for real-world society.
The current ABSA works, however, are mostly limited to the scenario of a single text piece, leaving the study in dialogue contexts unexplored.
To bridge the gap between fine-grained sentiment analysis and conversational opinion mining, in this work, we introduce a novel task of conversational aspect-based sentiment quadruple analysis, namely DiaASQ, aiming to detect the quadruple of \emph{target-aspect-opinion-sentiment} in a dialogue.
We manually construct a large-scale high-quality DiaASQ dataset in both Chinese and English languages.
We deliberately develop a neural model to benchmark the task, which advances in effectively performing end-to-end quadruple prediction, and manages to incorporate rich dialogue-specific and discourse feature representations for better cross-utterance quadruple extraction.
We hope the new benchmark will spur more advancements in the sentiment analysis community.
Our data and code are open at \url{https://github.com/unikcc/DiaASQ}.
\end{abstract}

\section{Introduction}
It is meaningful to empower machines to understand human opinion and sentiment, which motivates the study of sentiment analysis ~\cite{PangL07,mcdonald-etal-2007-structured,RenZZJ16,Cambria16}.
ABSA is an important branch of sentiment analysis aiming to detect the sentiment trends towards the fine-grained aspects of targets, which has received consistent research attention within last few years~\cite{LiBLLY18, fan-etal-2019-target, ChenLWZC20, Wu0RJL21, ChenZFLW22}.
The initial ABSA revolves around the study of \emph{aspect terms} and \emph{sentiment polarities}~\cite{tang-etal-2016-effective, FanFZ18,li2021unified}.
Later, the extraction of \emph{opinion terms} is considered, resulting in a triplet analysis (i.e., \emph{aspect-opinion-sentiment}) of ABSA~\cite{PengXBHLS20, ChenHLSJ21}.
The latest trend of ABSA has been upgraded into the quadruple form by adding the \emph{category} element into the triplet ABSA~\cite{CaiXY20, ZhangD0YBL21}.
The quadruple ABSA promisingly completes the ABSA definition and helps the comprehensive understanding of the opinion picture.

\begin{figure}[!t]
\vspace{10pt}
  \centering
  \includegraphics[width=0.95\columnwidth]{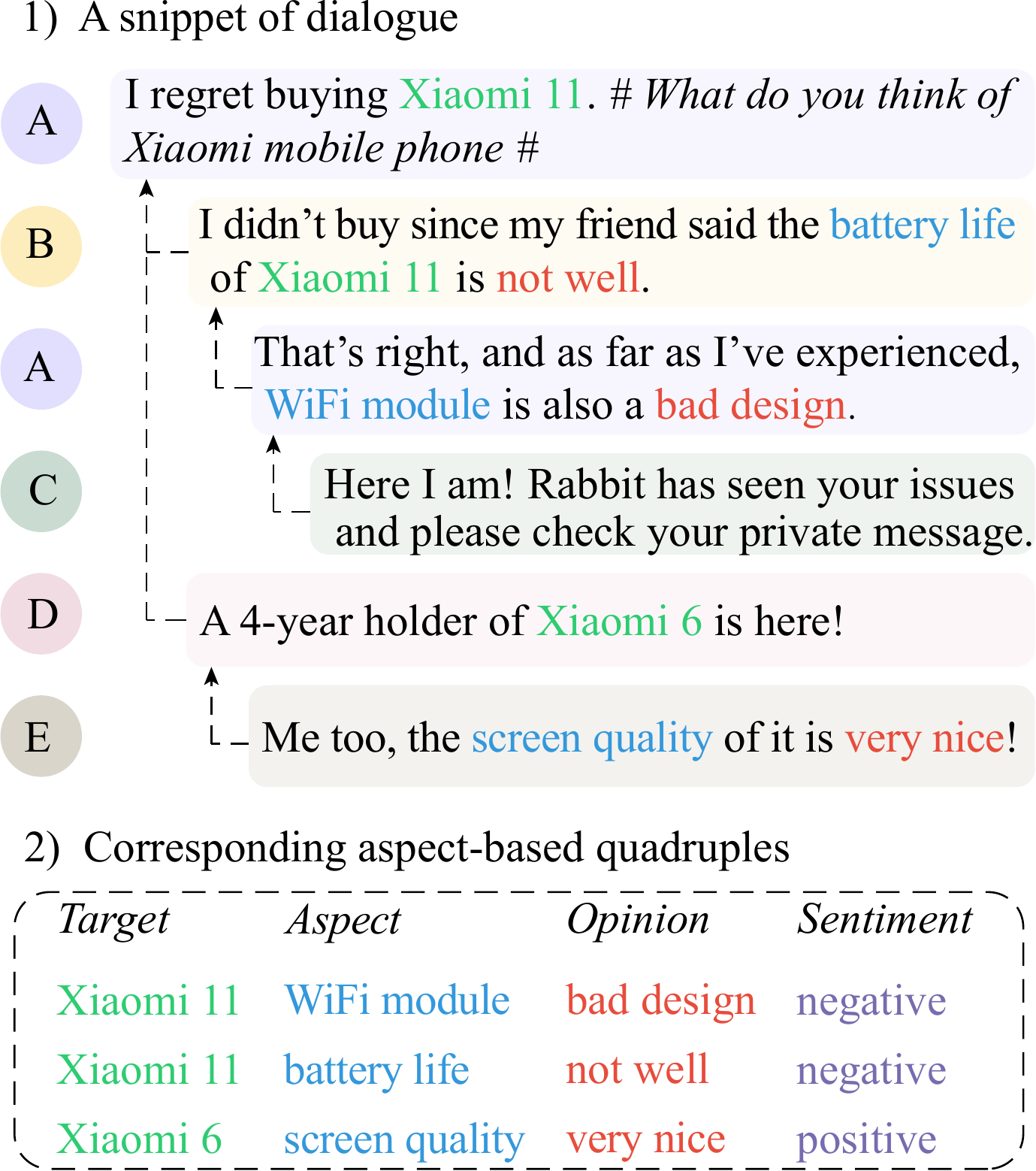}
  \caption{
  Illustration of the conversational aspect-based sentiment quadruple analysis (DiaASQ).
  The dialogue utterances produced by the corresponding speakers (marked at left) are organized into replying structure.
  }
  \label{fig:example}
  \vspace{-11pt}
\end{figure}

Yet we notice that all the current ABSA research is confined to the scenario of a single piece of text (i.e., sentence or document).
For example, currently the most popular ABSA benchmark, SemEval~\cite{PontikiGPPAM14, PontikiGPMA15, PontikiGPAMAAZQ16}, comes with only sentence-level annotations.
This may limit the application of ABSA.
Essentially, in the real-world environment ABSA has a broader application under dialogue contexts.
For example, people are more likely to discuss certain products, services, or politics on social media (e.g., Twitter, Facebook, Weibo) in the form of multi-turn and multi-party conversations.
Also, it is practically meaningful to develop sentiment-support dialog systems to facilitate the clinical diagnosis, and treatment~\cite{liu-etal-2021-towards}.
Unfortunately, no effort has been dedicated to the research of a holistic dialog-level ABSA.

In this paper, we consider filling the gap of dialogue-level ABSA.
We follow the line of recent quadruple ABSA and present a task of conversational aspect-based sentiment quadruple analysis, namely \textbf{DiaASQ}.
DiaASQ sets the goal to detect the fine-grained sentiment quadruple of \emph{target-aspect-opinion-sentiment} given a conversation text, i.e., an opinion of sentiment polarity has been expressed toward the target with respect to the aspect.
As exemplified in Fig.~\ref{fig:example}, multiple users (speakers) on social media discuss different angles of a product (i.e., `\emph{Xiaomi}' brand cellphone) in dialogue threads of multiple turns. 
The task aims to extract three quadruples over the dialog: 
(\emph{`Xiaomi 11', `WiFi module', `bad design', `negative'}),
(\emph{`Xiaomi 11', `battery life', `not well', `negative'})
and (\emph{`Xiaomi 6', `screen quality', `very nice', `positive'}).

To benchmark the task, we manually annotate a large-scale DiaASQ dataset.
We collect millions of conversational corpus of source comments and discussions closely related to electronic products from Chinese social media.
We hire well-trained workers to explicitly label the DiaASQ data (i.e., the elements of quadruples, targets, aspects, opinions, and sentiments) based on the crowd-sourcing technique, which ensures a high quality of annotations.
Finally, we yield the dataset with 1,000 dialogue snippets in total with 7,452 utterances.
To facilitate the multilinguality of the benchmark, we further translate and project the annotations into English.
Data statistics show that each dialog involves around 5 speakers, and 22.2\% of the quadruples are in the cross-utterance format.

Compared with previous single-text-based ABSA, DiaASQ challenges in two main aspects.
First, DiaASQ includes four subtasks.
Directly applying the existing best-performing graph-based ABSA model to enumerate all possible target, aspect, and opinion terms could cause a combinatorial explosion.
Second, the elements of a quadruple are scattered around the whole conversation due to the complex replying structure, which 
requires the model to do cross-utterance extraction.
To solve these challenges, we present an end-to-end DiaASQ framework.
Specifically, based on the grid-filling method~\cite{wu-etal-2020-grid}, we re-design the tagging scheme to fulfill the four subtasks in one shot effectively.
Moreover, during the dialogue text encoding, we additionally model the dialogue-specific representations for utterance interaction and meanwhile encode the relative distance as cross-utterance features. 
Experiments on the DiaASQ data indicate that our model shows significant superiority than several strong baselines.

To sum up, this work contributes in threefold:
\begin{compactitem}
  \item We pioneer the research of dialogue-level aspect-based sentiment analysis.
  Specifically, we introduce a conversational aspect-based sentiment quadruple analysis (DiaASQ) task. 
  \item We release a dataset for the DiaASQ task in both Chinese and English languages, which is of high quality and at a large scale.
  \item We introduce a model to benchmark the DiaASQ task.
  Our method solves the task end-to-end and meanwhile effectively learns the dialogue-specific features for better cross-utterance sentiment quadruple extraction.
  
\end{compactitem}

\vspace{-1mm}
\section{Related Work}

\begin{table}[!t]
 \setlength{\tabcolsep}{0.8mm}
  \centering
\resizebox{1\columnwidth}{!}{
  \begin{tabular}{lccccc}
  \hline
  \textbf{} & ASTE & TOWE & MAMS & CASA & DiaASQ \\
  \hline
  Target    & \textcolor{ired}{\ding{55}} & \textcolor{ired}{\ding{55}} & \textcolor{ired}{\ding{55}} & \textcolor{igreen}{\ding{51}} & \textcolor{igreen}{\ding{51}} \\
  Aspect    & \textcolor{igreen}{\ding{51}} & \textcolor{igreen}{\ding{51}} & \textcolor{igreen}{\ding{51}} & \textcolor{ired}{\ding{55}} & \textcolor{igreen}{\ding{51}} \\
  Opinion   & \textcolor{igreen}{\ding{51}} & \textcolor{igreen}{\ding{51}} & \textcolor{igreen}{\ding{51}} & \textcolor{igreen}{\ding{51}} & \textcolor{igreen}{\ding{51}} \\
  Polarity & \textcolor{igreen}{\ding{51}} & \textcolor{ired}{\ding{55}} & \textcolor{igreen}{\ding{51}} & \textcolor{igreen}{\ding{51}} & \textcolor{igreen}{\ding{51}} \\
  \hdashline 
  Dialogue-level   & \textcolor{ired}{\ding{55}} & \textcolor{ired}{\ding{55}} & \textcolor{ired}{\ding{55}} & \textcolor{igreen}{\ding{51}} & \textcolor{igreen}{\ding{51}} \\
Multi-sentiment       & \textcolor{ired}{\ding{55}} & \textcolor{ired}{\ding{55}} & \textcolor{igreen}{\ding{51}} & \textcolor{ired}{\ding{55}} & \textcolor{igreen}{\ding{51}} \\
Multilingual     & \textcolor{ired}{\ding{55}} & \textcolor{ired}{\ding{55}} & \textcolor{ired}{\ding{55}} & \textcolor{ired}{\ding{55}} & \textcolor{igreen}{\ding{51}} \\
  \hline
  \end{tabular}
}
  \vspace{-3pt}
  \caption{\label{table_corpus_comparision}
A comparison between our DiaASQ dataset and existing popular ABSA datasets, including: ASTE \cite{PengXBHLS20}, TOWE \cite{fan-etal-2019-target}, MAMS \cite{jiang-etal-2019-challenge}, and CASA \cite{song-casa-2022}.
  }
  \vspace{-10pt}
  \end{table}

\vspace{-1.2mm}
\subsection{Fine-grained Sentiment Analysis}

\vspace{-1mm}
All the existing ABSA tasks and their derivations revolve around predicting several elements or combinations: \emph{aspect term}, \emph{sentiment polarity}, \emph{opinion term}, \emph{aspect category}\footnote{
For example, the aspect `\emph{WiFi module}' in Fig.~\ref{fig:example} belongs to the \textit{hardware} category).
}, \emph{target}.
The initial ABSA task aims to classify the sentiment polarities given aspects~\cite{tang-etal-2016-effective,FanFZ18,li2021unified}.
Later, a wide range of new compound ABSA-related tasks is proposed, such as aspect-opinion paired extraction~\cite{zhao-etal-2020-spanmlt, Wu0RJL21}, aspect-category prediction~\cite{wang-2019-aspect, jiang-etal-2019-challenge, dai-etal-2020-multi}, triplet extraction~\cite{PengXBHLS20, ChenHLSJ21, chen-2022-span}, and structured opinion extraction~\cite{ShiLL0J22,Wu0LZLTJ22}, etc.

The latest attention has been placed on the quadruple or quintuple ABSA, where the \emph{aspect category} element is added into the triplet extraction~\cite{CaiXY20,ZhangD0YBL21,liu-etal-2021-comparative,Feiijcai22UABSA}.
Compared to all prior ABSA tasks, the sentiment quadruples provide much more complete opinion details that can facilitate downstream applications better.
In this work, we follow this line,
while our work differs in three aspects.
First, we consider adding the element of \emph{target} instead of \emph{category}.
Second, current quadruple and quintuple ABSA datasets all are incrementally annotated based on the existing SemEval data~\cite{PontikiGPPAM14, PontikiGPMA15, PontikiGPAMAAZQ16}; while we newly craft our data from real-world environment.
Third, this work mainly focuses on the conversation contexts instead of sentence pieces.

\vspace{-3pt}
\subsection{Dialogue Opinion Mining}

\vspace{-1pt}
In NLP community, dialogue applications show increasing impacts to real-world environments~\cite{LiaoLZHC21,ni2022recent,LiaoTMYHC22}.
The emotion and sentiment analysis in conversation scenarios is an essential branch of opinion mining.
Previous dialogue-level opinion mining has been limited to the coarse granularity, where the representative task is dialogue emotion detection~\cite{li-etal-2020-modeling,hu-etal-2021-dialoguecrn,li-etal-2022-emocaps}.
Yet as we indicated earlier, sentiment analysis in conversation at a fine-grained level has practical value.
In this paper, we pioneer the research of dialogue-level ABSA, presenting the conversational aspect-based sentiment quadruple analysis task. 

In Table~\ref{table_corpus_comparision} we compare our DiaASQ data with existing popular ASBA benchmarks.
It is worth noticing that, although CASA~\cite{song-casa-2022} is a dialogue-level sentiment analysis dataset, it may fail to provide a comprehensive understanding of opinion status due to the absence of key elements (e.g., \textit{aspect}).
In contrast, our DiaASQ dataset covers \textit{target}, \textit{aspect}, \textit{opinion} and \textit{sentiment}, which is by now the most comprehensive ABSA benchmark among all the other corpus.
In addition, sentiment understanding in DiaASQ is more complex and thus more challenging.
For example, one aspect term could correspond to multiple sentiments.
Besides, DiaASQ contains both Chinese and English versions, which will facilitate the research community to different languages.

\begin{figure}[!t]
  \centering
  \includegraphics[width=0.6\columnwidth]{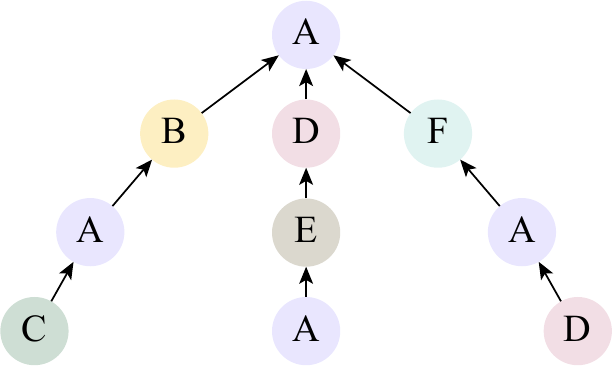}
  \vspace{-3pt}
  \caption{
  The tree-like dialogue replying structure.
}
  \label{fig:tree}
  \vspace{-10pt}
\end{figure}

\section{Data Construction}
\vspace{-1mm}
We construct a new dataset to facilitate the DiaASQ task.
The raw corpus is collected from the largest Chinese social media, Weibo\footnote{\url{https://weibo.com/}}.
We crawl nine million posts and comments from the tweets history of 100 verified digital bloggers.
Each conversation is derived from a root post, and multiple users (i.e., multiple speakers) are attended to reply to a predecessor post.
The multi-thread and multi-turn dialogue forms a tree structure, as illustrated in Fig.~\ref{fig:tree}.
We preprocess the raw dialogues to make the contexts integrated.
First, we filter the topic-related conversations by a manually created keyword dictionary in the mobile phone field, which includes hundreds of hot words, like phone band names, aspects words to describe a mobile phone, etc.
Then, we normalize the tweet language expressions (e.g., abusive language, hate speech) by human examination or consulting lexicons; we prune away those meaningless replying branches that deviate too much from the main topic.
We also limit the maximum number of utterances to ten for better controllable modeling.
After a strict cleaning procedure, we obtain the final 1,000 dialogues.

\begin{table*}[!t]
\fontsize{9.5}{11}\selectfont
    \centering
\resizebox{0.99\textwidth}{!}{
     \def\narrtablewidth{0.72 cm}
     \def\maintablewidth{0.77 cm}
     \def\widetablewidth{0.88 cm}
     \def\maxtablewidth{0.82 cm}
    \begin{tabular}
    {p{0.43 cm} p{0.68 cm}
    p{\narrtablewidth} p{\maintablewidth} p{\widetablewidth} 
    p{\narrtablewidth} p{\maintablewidth} p{\widetablewidth} 
    p{\maxtablewidth} p{\maintablewidth} p{\maxtablewidth} 
    p{\narrtablewidth} p{\maintablewidth} p{\widetablewidth}
    }
    \hline
     & & \multicolumn{3}{c}{\bf Dialogue} & & \bf Items & & \multicolumn{3}{c}{\bf Pairs} & \multicolumn{3}{c}{\bf Quadruples} \\
    \cmidrule(r){3-5}\cmidrule(r){6-8}\cmidrule(r){9-11} \cmidrule(r){12-14}
      & & Dia. & Utt. & Spk. & Tgt. & Asp. & Opi. & Pair$_{t\text{-}a}$ & Pair$_{t\text{-}o}$ & Pair$_{a\text{-}o}$ & Quad. & Intra. & Cross. \\
    \hline
    \multirow{4}{*}{\bf ZH}  &
    Total & 1,000 & 7,452 & 4,991 & 8,308 & 6,572 & 7,051 & 6,041 & 7,587 & 5,358 & 5,742 & 4,467 & 1,275\\
    \cdashline{2-14}
    & Train & 800   & 5,947 & 3,986 & 6,652 & 5,220 & 5,622 & 4,823 & 6,062 & 4,297 & 4,607 & 3,594 & 1,013\\
    & Valid & 100   & 748   & 502   & 823   & 662   & 724   & 621   & 758   & 538   & 577   & 440   & 137  \\
    & Test  & 100   & 757   & 503   & 833   & 690   & 705   & 597   & 767   & 523   & 558   & 433   & 125  \\
   \hline
   
   \hline
    \multirow{4}{*}{\bf EN}  &
Total & 1,000 & 7,452 & 4,991 & 8,264 & 6,434 & 6,933 & 5,894 & 7,432 & 4,994 & 5,514 & 4,287 & 1,227\\
    
    \cdashline{2-14}
    
& Train & 800   & 5,947 & 3,986 & 6,613 & 5,109 & 5,523 & 4,699 & 5,931 & 3,989 & 4,414 & 3,442 & 972  \\
& Valid & 100   & 748   & 502   & 822   & 644   & 719   & 603   & 750   & 509   & 555   & 423   & 132  \\
& Test  & 100   & 757   & 503   & 829   & 681   & 691   & 592   & 751   & 496   & 545   & 422   & 123  \\
    \hline
    \end{tabular}
    }
    \vspace{-2pt}
    \caption{
      Data statistics.
      `Dia.', `Utt.', and `Spk.' refer to dialogue, utterance, and speaker, respectively.
 `Tgt', `Asp', and `Opi' refer to target, aspect, and opinion terms, respectively.
 `Intra' and `Cross' refer to the intra-/cross-utterance quadruples.
 A quadruple is cross-utterance if any two elements of the (target, aspect, and opinion) in one quadruple distribute in different utterances.
 Since some words will be added, dropped, or merged during the translating process, the numbers of annotation items in Chinese and English versions of datasets are somewhat different.
    }
    \vspace{-5pt}
    \label{table_corpus_statistics}
\end{table*}

During the annotation stage, all the conversation texts are labeled with a team of crowd-workers who are pre-trained under the SemEval ABSA~\cite{PontikiGPPAM14} annotation guideline\footnote{
Appendix $\S$~\ref{Specification on Data Acquisition} details the data acquisition and annotation.
}.
Also, the linguistic and computer science experts inspect the labeling schema.
After annotating, annotators are required to cross-examine the labels.
Also, some automatic rules are applied to verify the labeling consistency.
Finally, Cohen’s Kappa score of quadruples is 0.86, which indicates our annotation corpus has reached a high-level agreement. 

\begin{figure}[!t]
  \centering
  \includegraphics[width=0.68\columnwidth]{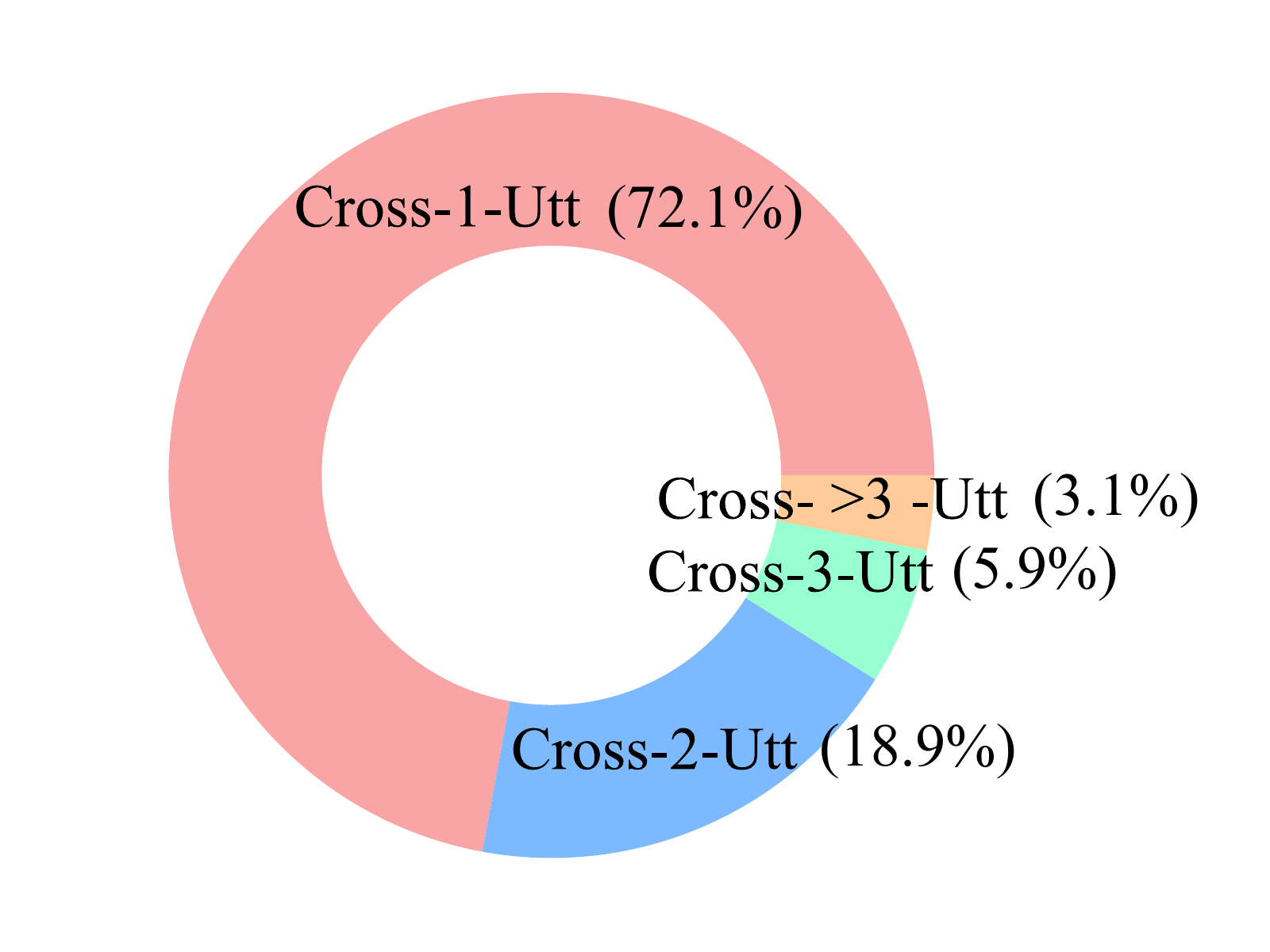}
  \caption{
  The ratio of cross-utterance quadruples.
  We define the max utterance-level distance between every two items in one quadruple as the number of cross-utterance.
  For example, the first quadruple in Fig.~\ref{fig:example} crosses two utterances.
  }
  \label{cross-utt}
  \vspace{-10pt}
\end{figure}

\vspace{2pt}
\textbf{Data Insights.}
We randomly split the conversation snippets into train/valid/test sets, in the ratio of 8:1:1.
The Chinese version of the dataset contains a total of 7,452 utterances, and 5,742 sentiment quadruples, while the English version contains 5,514 quadruples, which are far bigger numbers than the existing quadruple and quintuple ABSA datasets~\cite{CaiXY20, ZhangD0YBL21}.
Also, there is an average of one sentimental expression in each utterance.
Such annotation density makes it quite convenient for task prediction.
The data statistics are shown in Table~\ref{table_corpus_statistics}.
Each dialog has around five speakers on average, and the dataset contains 1,275 (22.2\%, in Chinese) and 1,227 (22.3\%, in English) cross-utterance quadruples, respectively.
In Fig.~\ref{cross-utt}, we show the ratio of quadruples of the dataset under different cross-utterance levels.
More data statistics are shown in Appendix $\S$~\ref{Extended Data Specification}.

\section{Grid-tagging Task Modeling with Renewed Label Scheme}
\label{Grid-tagging Modeling with Renewed Label Scheme}
The input of the DiaASQ task includes a dialogue $D=\{u_1,\cdots,u_n\}$ with the corresponding replying record $l=\{l_1,\cdots,l_n\}$ of utterances, where $l_i$ denotes that $i$-th utterance replies to $l_i$-th utterance.

Each  $u_i=\{w_{1}, \cdots,w_{m}\}$ denotes $i$-th utterance text and $m$ is the length of utterance $u_i$.
The replying record $l$ reflects the hierarchical tree structure of $D$.
Based on the input $D$ and $l$, DiaASQ aims to extract all possible (\emph{target}, \emph{aspect}, \emph{opinion}, \emph{sentiment}) quadruples, denoted as $Q=\{t,a,o,p\}_{k=1}^K$.
The \emph{target}, \emph{aspect} or \emph{opinion} term ($t_k, a_k, o_k$) is a sub-string of an utterance text $u_i$.
The sentiment $p_k$ is a category label $\in\{pos,neg,other\}$.

DiaASQ naturally includes four subtasks.
Different popular end-to-end ABSA systems can be utilized to solve our DiaASQ, such as the graph-based~\cite{zhu-tal-01-clser,ChenZFLW22}, seq-to-seq~\cite{Zhang0DBL20,mu-etal-21-pte} and grid-tagging models~\cite{wu-etal-2020-grid}.
Yet enumerating all possible terms with graph-based methods will cost computational efficiency, while seq-to-seq methods suffer from exposure bias.
The grid-tagging method advances in higher efficiency, i.e., $\mathcal{O}(n^2)$ complexity, where $n$ denotes the sequence length.
However, the labeling scheme in~\cite{CaiXY20, ZhangD0YBL21} only supports term-pair extraction (i.e., \emph{aspect} and \emph{opinion} terms), which fails to directly solve our DiaASQ that requires term-triple extraction (i.e., \emph{target}, \emph{aspect} and \emph{opinion} terms).
Here we inherit the success of the grid-tagging method for an end-to-end solution and re-design the labeling scheme to fit our needs.

\begin{figure}[!t]
  \centering
  \includegraphics[width=0.95\columnwidth]{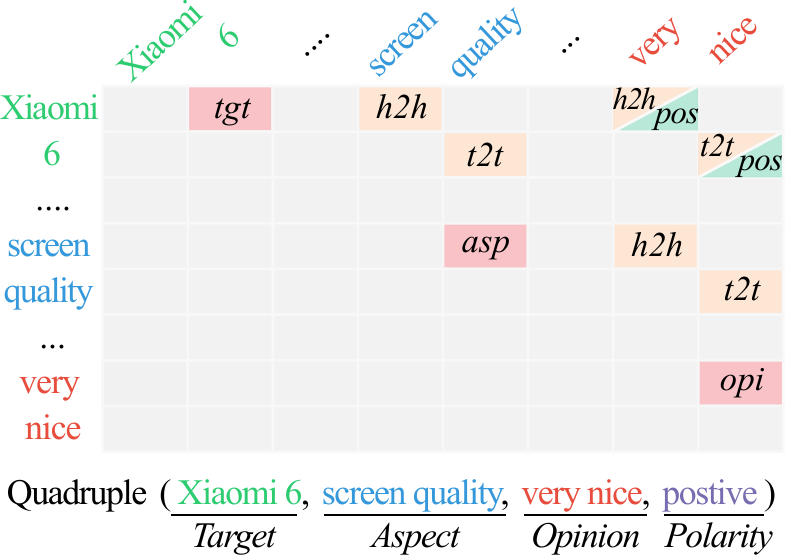}
  \caption{Tagging scheme for quadruple extraction. }
  \label{fig:matrix}
  \vspace{-4mm}
\end{figure}

To reach the goal, we re-decompose the task into three joint jobs: detections of the entity boundary, entity pair, and sentiment polarity.
We renew the labeling scheme of grid-tagging in support of these jobs, which is shown in Fig.~\ref{fig:matrix}.

\vspace{2pt}
$\bullet$ \textbf{Entity Boundary Labels}:
We use \textit{tgt, asp, opi} to denote the token-level relations between the head and tail of a \emph{target}, \emph{aspect}, and \emph{opinion} term, respectively.
For example, the \textit{tgt} between `\emph{Xiaomi}' and `\emph{6}' denotes a \emph{target} term of `\emph{Xiaomi 6}'.

\vspace{2pt}
$\bullet$ \textbf{Entity Pair Labels}:
We then need to link different types of terms together as a combination.
To represent the relation between entities, we devise two labels: \emph{h2h} and \emph{t2t}, both of which align the head and tail tokens between a pair of entities in two types.
For example, the head words of `\emph{Xiaomi}' (\emph{target}) and `\emph{screen}' (\emph{aspect}) is connected with \emph{h2h}, while the tail words of `\emph{6}' (\emph{target}) and `\emph{quality}' (\emph{aspect}) is connected with \emph{t2t}.
By labeling a chain of term pairs in different types, we form a triplet of ($t_k, a_k, o_k$).

\vspace{2pt}
$\bullet$ \textbf{Sentiment Polarity Labels}:
By adding a sentiment category label $p_k$, we then form a quad $q_k$=($t_k, a_k, o_k, p_k$).
Since the \emph{target} and \emph{opinion} terms together determine a unique sentiment, we assign the category label between the heads and tails of these two terms, as shown in Fig.~\ref{fig:matrix}.

\begin{figure*}[!t]
\centering
\includegraphics[width=0.86\textwidth]{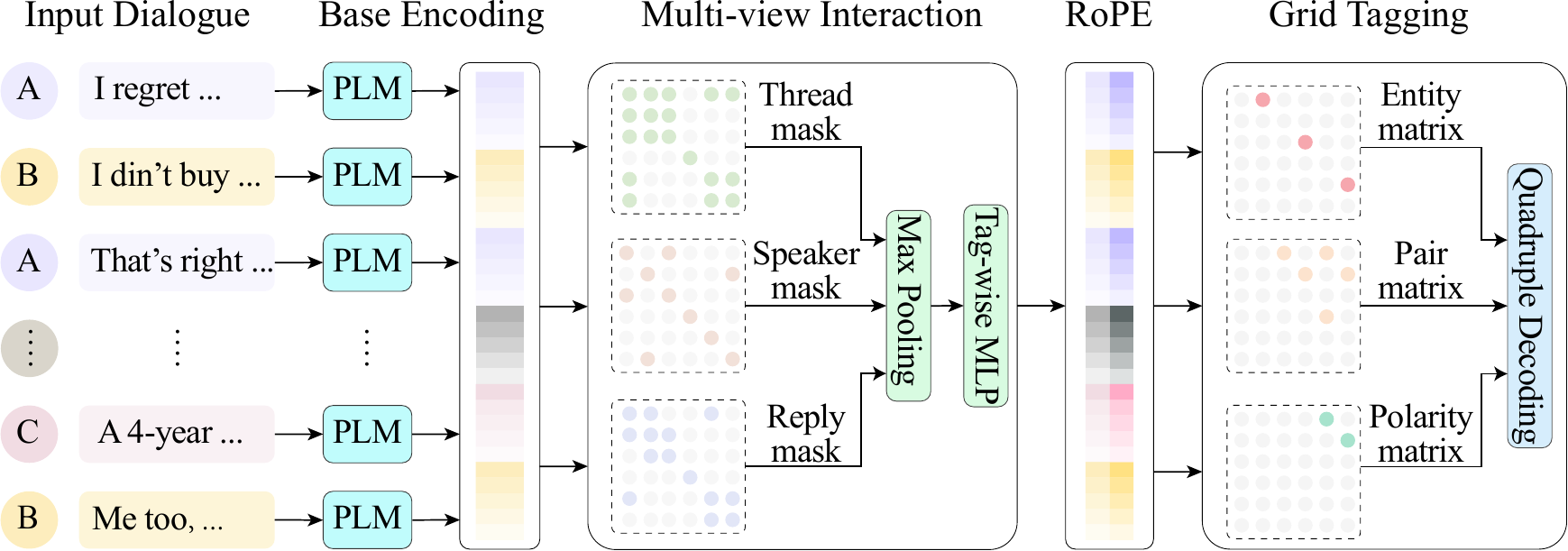}
\caption{
The overall framework of our DiaASQ model.
First, the base encoder learns base contextual representations for the input dialogue texts.
The multi-view interaction layer then aggregates dialogue-specific feature representations, such as the threads, speakers, and replying information.
We further fuse the Rotary Position Embedding (RoPE), where the relative dialogue distance information helps guide better discourse understanding. 
Finally, the system decodes all the quadruples based on the grid-tagging labels.
}
\label{fig:model}
\vspace{-5pt}
\end{figure*}

\section{DiaASQ Model}
We present a DiaASQ model to accomplish the task based on the above grid-tagging label scheme.
Fig.~\ref{fig:model} shows the overall architecture.

\subsection{Base Encoding}
We adopt a pre-trained language model (PLM), e.g., BERT~\cite{DevlinCLT19}, to encode the dialogue utterances.
However, the length of a whole dialogue may far exceed the max length that BERT can accept.
We thus encode each utterance with a separate PLM one by one.
We use the [CLS] and [SEP] tokens to separate each utterance $u_i$.
\setlength\abovedisplayskip{2pt}
\setlength\belowdisplayskip{2pt}
\begin{align}
&u^{'}_i = < \text{[CLS]}, w_{1}, \cdots , w_{m}, \text{[SEP]} > \,, \\
\label{base-rep} &\bm{H_i} = \bm{h}_{cls},  \bm{h}_1, \cdots, \bm{h}_m, \bm{h}_{sep} = \text{PLM}( u^{'}_i ) \,,
\end{align}
where $\bm{h}_m$ is the contextual representation of $w_{m}$.

\vspace{-5pt}
\subsection{Dialogue-specific Multi-view Interaction}

To strengthen the awareness of the dialogue discourse, we then introduce a multi-view interaction layer to learn the dialogue-specific features.
This layer is built upon the multi-head self-attention~\cite{VaswaniSPUJGKP17}.
Inspired by~\cite{ShenCQX21, ZhaoZ0LJW022}, we use three types of features: dialogue threads, speakers, and replying.
Specifically, we realize the idea by constructing attention masks $\bm{M}^c$ that carry the bias of such prior features, controlling the interactions between tokens.
And $c\in\{Th, Sp, Rp\}$ represents different types of token interaction, i.e., thread, speaker, and replying, respectively. 
\begin{equation}
\begin{aligned}
    \bm{H}^c &= \text{Masked-Att}(\bm{Q},\bm{K},\bm{V},\bm{M}^c)  \\
    &= \text{Softmax} (\frac{(\bm{Q}^T \cdot  \bm{K}) \odot \bm{M}^c }{\sqrt{d}} ) \cdot \bm{V} \,,
\end{aligned}
\end{equation}
where $\bm{Q}$=$\bm{K}$=$\bm{V}$=$\bm{H}\in\mathbb{R}^{N\times d}$ is the representation of the whole dialogue sequence obtained by concatenating token representations of each utterance ($\bm{H}_i$ in Eq.~\eqref{base-rep}), $N$ is the token-level length of $D$,
and $\odot$ is element-wise production.
The value of $\bm{M}^c \in \mathbb{R}^{N\times N}$ is defined as follows:

\vspace{2pt}
$\bullet$ \textbf{Thread Mask}: $\bm{M}^{Th}_{ij}$=1 if the $i^{th}$ and $j^th$ token belong to the same dialogue thread.

\vspace{2pt}
$\bullet$ \textbf{Speaker Mask}: $\bm{M}^{Sp}_{ij}$=1 if the $i^{th}$ and $j^{th}$ token are derived from the same speaker.

\vspace{2pt}
$\bullet$ \textbf{Reply Mask}: $\bm{M}^{Rp}_{ij}$=1 if the two utterances containing the $i^{th}$ and $j^{th}$ token respectively have a replying relation.

We then conduct Max-Pooling over the masked representations, followed by a tag-wise MLP layer to yield the final feature representation $\bm{v}^c_i$:
\begin{align}
    \bm{H}^f = \text{Max-Pooling}&(\bm{H}^{Th}, \bm{H}^{Sp}, \bm{H}^{Rp}) \,, \\
 \label{eq_dense}  \bm{v}^r_i = &\text{MLP}^r(\bm{h}^f_i) \,,
\end{align}
where $r\in\{tgt,\cdots,h2h,\cdots,pos,\cdots,\epsilon_{ent},\cdots\}$ indicates a specific label, and $\epsilon_{ent}$ denotes the non-relation label in the entity boundary matrix.

\vspace{-2pt}
\subsection{Integrating Dialogue Relative Distance}

Limited by the PLM, we can only encode each utterance separately, potentially hurting the conversational discourse.
To compensate for it, we consider fusing the Rotary Position Embedding (RoPE)~\cite{abs-2104-09864} into token representations.
RoPE dynamically encodes the relative distance globally between utterances at the dialogue level.
Introducing such distance information can help guide better discourse understanding. 
\begin{equation}
  \bm{u}^r_i = \boldsymbol{\mathcal{R}}(\theta, i) \bm{v}^r_i \,,
  \label{eq.rope}
\end{equation}
where 
$\boldsymbol{\mathcal{R}}(\theta, i)$ is a positioning matrix parameterized by $\theta$ and the absolute index $i$ of $\bm{v}^r_i$.

\vspace{-2pt}
\subsection{Quadruple Decoding}

Based on each tag-wise representation $\bm{u}^r_i$, we finally calculate the unary score between any token pair in terms of label $r$:
\begin{equation}
  s^r_{ij} = (\bm{u}^r_i)^T \bm{u}^r_j  \,,
  \label{eq_biaffine}
\end{equation}
where $s^r_{ij}$ is the probability that the relation label between $w_i$ and $w_j$ is $r$.
Then we put a softmax layer over all elements in each matrix to determine the relation label $r$.
For example, the probability of entity boundary matrix can be obtained via:
\begin{equation}
\label{eq-3} p^{ent}_{ij} = \text{Softmax}([s^{\epsilon_{ent}}_{ij}; s^{tgt}_{ij}; s^{asp}_{ij}; s^{opi}_{ij}]) \,.
\end{equation}

Obtaining all the labels in the grid, we decode all the quadruples based on the rules stated in $\S$~\ref{Grid-tagging Modeling with Renewed Label Scheme}.
  
\begin{table*}[!t]
\fontsize{9.5}{11}\selectfont
    \centering
     \def\maintablewidth{1.0}
    \begin{tabular}{p{0.5cm} p{3.3cm} p{\maintablewidth cm}<{\centering} p{\maintablewidth cm}<{\centering} p{\maintablewidth cm}<{\centering} p{\maintablewidth cm}<{\centering} p{\maintablewidth cm}<{\centering} p{\maintablewidth cm}<{\centering} p{\maintablewidth cm}<{\centering} p{\maintablewidth cm}}
    \hline
     & & \multicolumn{3}{c}{Span Match (F1)} & \multicolumn{3}{c}{Pair Extraction (F1)} & \multicolumn{2}{c}{Quadruple (F1)} \\
    \cmidrule(r){3-5}\cmidrule(r){6-8}\cmidrule(r){9-10}
      & & T & A & O & T-A & T-O & A-O & Micro & Iden. \\
    \hline
    \multirow{7}{*}{ZH}  &
    CRF-Extract-Classify & \textbf{91.11} & 75.24 & 50.06 & 32.47 & 26.78 & 18.90 & 8.81 & 9.25 \\
     & SpERT &  90.69 & 76.81 & 54.06 & 38.05 & 31.28 & 21.89 & 13.00 & 14.19 \\
    & ParaPhrase & / & / & / & 37.81 & 34.32 & 27.76 & 23.27 & 27.98 \\
    & Span-ASTE & / & / & / & 44.13 & 34.46 & 32.21  & 27.42  & 30.85 \\
   & \quad w/o PLM & / & / & / & 28.36 & 24.81 & 22.50  & 8.96  &11.21 \\
   \cdashline{2-10}
    & Ours & 90.23 & \textbf{76.94} & \textbf{59.35} & \bf 48.61 & \bf 43.31 & \bf 45.44 & \textbf{34.94} & \textbf{37.51} \\
   & \quad w/o PLM & 85.52 & 75.21 & 47.15 & 34.72& 26.16 & 30.73 & 14.21 & 17.55  \\
   \hline
   
   \hline
       \multirow{7}{*}{EN}  &
    CRF-Extract-Classify & 88.31 & 71.71 & 47.90 & 34.31 & 20.94 & 19.21 & 11.59 & 12.80 \\
     & SpERT &  87.82 & 74.65 & 54.17 & 28.33 & 21.39 & 23.64 & 13.07 & 13.38 \\
    & ParaPhrase & / & / & / & 37.22 & 32.19 & 30.78 & 24.54 & 26.76 \\
    & Span-ASTE & / & / & / & 42.19 & 30.44 & \bf 45.90  & 26.99  & 28.34 \\
   & \quad w/o PLM & / & / & / & 27.26 & 20.63 & 44.62  & 13.84  & 14.17 \\
   \cdashline{2-10}
    & Ours & \bf 88.62 & \bf 74.71 & \bf 60.22 & \bf 47.91 & \bf 45.58 & 44.27 & \bf 33.31 & \bf 36.80 \\
    & \quad w/o PLM & 83.02 & 68.89 & 53.87 & 32.53 & 31.09 & 35.59 & 15.68 & 19.57\\
   \hline
    \end{tabular}
    \caption{
      Main results of the DiaASQ task.
      `T/A/O' represent Target/Aspect/Opinion, respectively.
      All the scores are averaged values over five runs under different random seeds.
      Since ParaPhrase and Span-ASTE do not distinguish the term types, we here do not measure the performances of span match.
      Note that `w/o PLM' indicates that we use randomly initialized word2vec to encode the text.
    }
    \label{table_main}
    \vspace{-4mm}
\end{table*}

\vspace{-2pt}
\subsection{Learning}

The training target is to minimize the cross-entropy loss of each subtask:
\begin{equation}
  \mathcal{L}_k = -\frac{1}{G\cdot N^2} \sum_{g=1}^G \sum_{i=1}^{N} \sum_{j=1}^{N} \bm{\alpha}^k \, y^k_{ij}\log (p^k_{ij}) \,,
\end{equation}
where $k\in\{ent,pair,pol\}$ indicates a subtask,
$N$ is the total token length in a dialogue, and $G$ is the total training data instances.
$y^k_{ij}$ is ground-truth label,
$p^k_{ij}$ is the prediction.
The label types (stated in Section ~\ref{Grid-tagging Modeling with Renewed Label Scheme}) are imbalanced.
Thus we apply a tag-wise weighting vector $\bm{\alpha}^k$ to counteract this.
We then add up all three loss items as the final one:
\begin{equation}
\mathcal{L} = \mathcal{L}_{ent} + \beta \mathcal{L}_{pair} + \eta \mathcal{L}_{pol}  \,.
\end{equation}

\section{Experiment}

\vspace{-2mm}
\subsection{Settings}

\vspace{-1mm}
We conduct experiments on our DiaASQ dataset to evaluate the efficacy of our proposed model.
We mainly measure the performances in terms of three angles:
1) \emph{span match}: the boundary of three types of term spans;
2) \emph{pair extraction}: the detection of span pair, i.e., \emph{Target}-\emph{Aspect}, \emph{Aspect}-\emph{Opinion} and \emph{Target}-\emph{Opinion};
3) \emph{quadruple extraction}: recognizing the full quad of DiaSAQ task.
We use the \emph{exact F1} as the metric: for span, a correct prediction should match both the left and right boundaries; for pair, match both two spans and the relation; for quad, match all four elements exactly.
The performance of quadruple extraction is our main focus.
We thus take the \emph{micro F1} and \emph{identification F1} respectively for measurements, where the micro F1 measures the whole quad, including the sentiment polarity.
In contrast, \emph{identification F1}~\cite{barnes-etal-2021-structured} does not distinguish the polarity.

We take the Chinese-Roberta-wwm-base~\cite{CuiCLQY21} and Roberta-Large~\cite{roberta-liu} as our base encoders for the Chinese and English datasets, respectively.
We put a 0.2 dropout rate on the BERT output representations.
MLP in Eq.~\eqref{eq_dense} has a 64-d hidden size.
The testing results are given by the models tuned on the developing set.
All experiments take five different random seeds, and the final scores are averaged over five runs.

As no prior method is deliberately designed for DiaASQ, we consider re-implementing several strong-performing systems closely related to the task as our baselines, including 
\textbf{CRF-Extract-Classify}~\cite{CaiXY20},
\textbf{SpERT}~\cite{EbertsU20}
\textbf{Span-ASTE}~\cite{XuCB20} and
\textbf{ParaPhrase}~\cite{ZhangD0YBL21}.
All baselines take the same PLM as used in our model except that \textbf{ParaPhrase} uses mT5-base~\cite{XueCRKASBR21}.

\subsection{Main Comparisons}

\begin{table*}[t]
  \centering
  \def\curwidth{1.68cm}
  \resizebox{2\columnwidth}{!}{

  \begin{tabular}{p{2.9 cm} p{\curwidth}<{\centering} p{\curwidth}<{\centering} p{\curwidth}<{\centering} p{\curwidth}<{\centering} p{\curwidth}<{\centering} p{\curwidth}<{\centering}}
  \hline
   & \multicolumn{3}{c}{ZH} & \multicolumn{3}{c}{EN} \\
  \cmidrule(r){2-4}\cmidrule(r){5-7} 
    & Overall& Intra-Utt. & Inter-Utt. & Overall & Intra-Utt. & Inter-Utt. \\
  \hline

  Ours & 34.94  & 37.95 & 23.21 & 33.31 & 37.65 & 15.76 \\
  \hline
  w/o All-Interaction & 34.04\scriptsize{($\downarrow$0.90)} & 37.40\scriptsize{($\downarrow$0.55)} &  20.95\scriptsize{($\downarrow$2.26)} & 32.51 \scriptsize{($\downarrow$0.80)}  & 37.23 \scriptsize{($\downarrow$0.32)} &  12.98 \scriptsize{($\downarrow$2.78)}\\
  \quad w/o Speaker & 34.43\scriptsize{($\downarrow$0.51)} & 37.82\scriptsize{($\downarrow$0.13}) & 21.90\scriptsize{($\downarrow$1.31}) & 33.06  \scriptsize{($\downarrow$0.25)}  &   37.68 \scriptsize{($\uparrow$0.03)}     & 14.20  \scriptsize{($\downarrow$1.56)}  \\
  \quad w/o Thread & 34.52\scriptsize{($\downarrow$0.42}) & 37.61\scriptsize{($\downarrow$0.34}) & 22.62\scriptsize{($\downarrow$0.59}) & 33.09 \scriptsize{($\downarrow$0.22)} &  37.33   \scriptsize{($\downarrow$0.32)}  & 15.09  \scriptsize{($\downarrow$0.67)} \\
 \quad w/o Reply & 34.26\scriptsize{($\downarrow$0.68}) & 37.06\scriptsize{($\downarrow$0.89}) & 22.91\scriptsize{($\downarrow$0.30}) & 32.82  \scriptsize{($\downarrow$0.49)} &    37.46  \scriptsize{($\downarrow$0.21)} & 13.50  \scriptsize{($\downarrow$2.26)} \\

\hdashline
w/o RoPE & 33.10\scriptsize{($\downarrow$1.84}) & 36.42\scriptsize{($\downarrow$1.53}) & 20.22\scriptsize{($\downarrow$2.99})  
         & 31.59  \scriptsize{($\downarrow$1.72)} & 36.44   \scriptsize{($\downarrow$1.21)}  & 12.22  \scriptsize{($\downarrow$3.54)} \\

\hdashline
w/o Lab.Wei. ($\bm{\alpha}^k$) & 33.52\scriptsize{($\downarrow$1.42}) & 36.63\scriptsize{($\downarrow$1.32}) & 20.93\scriptsize{($\downarrow$2.28}) & 32.54  \scriptsize{($\downarrow$0.77)} & 37.06  \scriptsize{($\downarrow$0.59)}  & 13.50   \scriptsize{($\downarrow$2.26)} \\

\hline
\end{tabular}}
  \caption{
    Ablation results (Micro F1).
    `w/o All-Interaction': removing all three multi-view interaction items.
  }
  \label{table_ablation}
    \vspace{-2mm}
\end{table*}

Table~\ref{table_main} compares the performances of different models on the DiaASQ task. 
We see that our proposed method achieves the overall best results under almost all measurements.
Besides, we have the following observations.

First, the performance divergences of different models on span detection are not significant, and all the methods perform well on the subtask.
We think this is mainly because, without considering the inter-relation between each type of term (T/A/O), recognizing the mentions is a pretty simple task.

Second, it is clear that our model starts surpassing the baselines on pair-wise detection.
Our system outperforms the second-best models over average 9\% of F1 score on almost all cases, i.e., T-A, T-O, and A-O.
This result verifies that our model is more effective than baselines on sentiment information extraction under the conversational scenario.
One exception is that the Span-ASTE slightly exceeds our model on A-O pair extraction in the English version dataset.
The possible reason is that aspect and opinion pair usually co-occur closely, and it has been a classical task for which span-aste can achieve competitive results. 

Finally and most importantly, our system shows huge wins on the quadruple extraction, with 7.52\% micro F1(=34.94-27.42) and 6.66\% identification F1(=37.51-30.85) improvements on the Chinese dataset, with 6.32\% micro F1(=33.31-26.99) and 8.46\% identification F1(=36.80-28.34) improvements on the English dataset, respectively.
This result evidently shows our model's efficacy on the task.
We also find that stripping off the PLMs hurts the task performances very prominently, even for the strong models.

\subsection{Ablation Study}

We now take a further step, examining the efficacy of several key designs in our method, including the dialogue-specific multi-view interaction, the relative distance embedding (RoPE), and the label-wise weighting mechanism.
The ablating results are shown in Table~\ref{table_ablation}.

First, we see that the different type of dialogue-specific interaction shows the varying influence.
For example, thread features show the overall most negligible impacts, which improve the F1 score of Inter-Utt by no more than 1\% in the two datasets.
In contrast, the speaker-aware and reply-aware interactions are more important that improve the score Inter-Utt by more than 1\%.
Interestingly, some ablations increase the performances in the intra-utterance case but decrease rapidly in the cross-utterance case.

Then, we witness the most significant performance drops when removing the RoPE feature.
Significantly, the F1 score of cross-utterance drops 2.99\% and 3.54\% in the Chinese and English datasets, respectively.
This result demonstrates the importance of modeling dialogue-level discourse information.
Finally, we see that the label-wise weighting mechanism used for task learning is also much crucial.
This finding is reasonable because the labels of different types in the grid among the whole dialogue are imbalanced and sparse, e.g., the positive tags are far less than the negative ones (i.e., $\epsilon_{ent}$).
Label-wise weighting helps effectively solve the label imbalance issue.

\begin{figure}[!t]
  \centering
  \begin{tikzpicture}
    \begin{axis}[
      ytick = {0,10,20,30,40},
      ymax=44,ymin=-3,
      ylabel=Quadruple (Micro F1),
      ylabel style = {yshift=-0.5em,font=\small},
      y tick label style = {font=\small\bfseries},
        height=0.53\columnwidth,
		width=1.05\columnwidth,
      legend style={at={(0.97, 1.39)}, draw=none, font=\small,
        /tikz/every even column/.append style={column sep=0.3cm},
        /tikz/every odd column/.append style={column sep=0.05cm,}},    
      legend image post style={xscale=1.0},
      legend columns=2,
      xtick = {0.1,0.2,0.3,0.4,0.5},
      xticklabels = {Intra-Utt, Cross-1-Utt, Cross-2-Utt, Cross-$\ge$3-Utt},
      xmax=0.45,xmin=0.05,
      x tick label style = {font=\small,yshift=2pt,xshift=-2pt,rotate=30},
      ]
      
  \draw [dashed] (0,10) -- (0.5,10);
  \addplot[thick, smooth, red, mark = *, mark size = 1pt, mark options={fill=red}] coordinates
  {
  (0.1,	38.47)
  (0.2,	29.32)
  (0.3,	19.05)
  (0.4,	15.38)
  };
  
  \addplot[thick, smooth,  densely dashed, orange, mark=*, mark size = 1pt, mark options={fill=orange}] coordinates
      {(0.1,37.59) (0.2,28.57) (0.3,9.68) (0.4,0) };
  \addplot[thick,smooth,  densely dashed, cyan, mark = *, mark size = 1pt, mark options={fill=cyan}] coordinates
  { (0.1,	37.71) (0.2,	22.86) (0.3,	14.81) (0.4,	10.34) };
 \addplot[thick, smooth, densely dashed, blue, mark = *, mark size = 0.5pt, mark options={fill=blue}] coordinates
  { (0.1,	32.29) (0.2,	12.61) (0.3,	10.00) (0.4,	0.00) };
  \legend{Ours, Ours w/o All-Interaction, Ours w/o RoPE, Span-ASTE}
  \end{axis}
    \end{tikzpicture}
  \vspace{-4pt}
  \caption{
  results on different cross-utterance levels.
  }
  \label{fig:cross}
  \vspace{-1em}
  \end{figure}
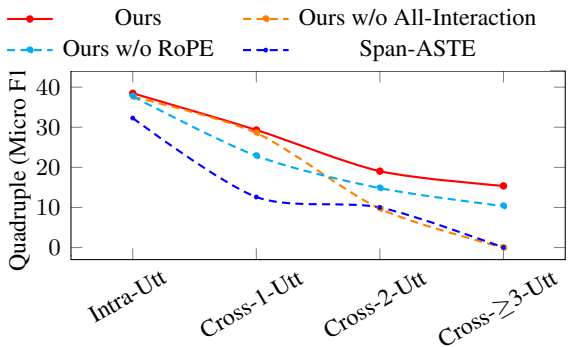

\vspace{-1mm}
\subsection{Further Analysis}

\vspace{-1mm}
In this section, we consider diving into the model performances and carry on an in-depth analysis to better understand the strengths of our method.

\vspace{2pt}
\noindent\textbf{Cross-utterance Quadruple Extraction.}
Earlier in Table~\ref{table_main}, we verify the superiority of our model.
We mainly credit its capability to effectively model the cross-utterance features.
Here we directly examine this attribute by observing the performances under different levels of the cross-utterance quad extraction.
As plotted in Fig.~\ref{fig:cross}, we observe the patterns that the more utterances quadruple across, the lower the performances all models can achieve.
Especially when the cross-utterance level $\ge$3, the baseline systems fail to recognize any single quad.
Nevertheless, our system can still well resolve the challenge, even in case of cross-$\ge$3-utterance.
Also, by comparing two of our ablated models, we learn that the dialogue-specific interaction features are more beneficial for handling the super-long-distance cross-utterance.
But the RoPE that carries discourse information contributes more to the short-range case (i.e., cross-1-utterance).

\vspace{2pt}
\noindent\textbf{Impact of Dialogue-level Distance Encoding.}
We equip our framework with dialogue-level relative distance embeddings (i.e., RoPE, a dynamic positioning feature), so as to enhance conversational discourse understanding.
Here we study the influence of using different dialogue-level distance embeddings.
We consider two other alternative solutions:
1) \emph{Relative position encoding}, which is a type of dense embedding of relative distances of utterance;
We directly add the embedding to the token relation probability vector in Eq.~\eqref{eq-3} to introduce this information. 
2) \emph{Global position encoding}, which is an absolute position embedding of the token. 
We utilize the global position by adding to the token representation $\bm{v}^r_i$ in Eq.~\eqref{eq_dense}.

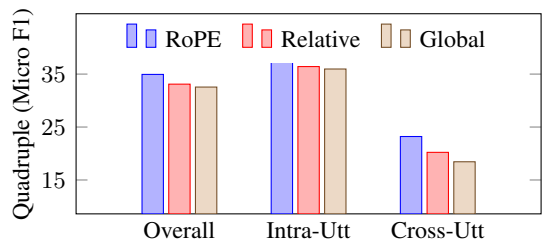
\begin{figure}[!t]
    \centering
    \begin{tikzpicture}[font=\small]
	\begin{axis}[
        height=0.55\columnwidth,
		width=1\columnwidth,
        ybar,
        ytick = {15,25,35},
        ymin=17, 
        ymax=38,
        bar width=8pt,
        ylabel shift=-5pt, 
        enlargelimits=0.4,
        legend style={at={(0.5,0.98)}, anchor=north, legend columns=-1,  draw=none,
        /tikz/every even column/.append style={column sep=0.2cm},
        /tikz/every odd column/.append style={column sep=0.05cm,}},
        legend image post style={xscale=1.2},
        legend columns=3,
        xtick style={draw=none},
        xticklabel style={yshift=5pt},
        ylabel=Quadruple (Micro F1),
        ylabel style = {yshift=-0.0em,font=\small},
        y tick label style = {font=\small\bfseries},
        symbolic x coords={Overall, Intra-Utt, Cross-Utt},
        xtick=data
		]
        \addplot coordinates {(Overall, 34.94) (Intra-Utt, 37.95) (Cross-Utt, 23.21)};
        \addplot coordinates {(Overall, 33.10) (Intra-Utt, 36.42) (Cross-Utt, 20.22)};
        \addplot coordinates {(Overall, 32.55) (Intra-Utt, 35.96) (Cross-Utt, 18.44)};
 	\legend{RoPE, Relative, Global}
	\end{axis}
\end{tikzpicture}
    \caption{
    Influences of using difference distance-encoding methods.
    }
    \label{fig:distance}
    \vspace{-4mm}
\end{figure}

We study the performance changes on quadruple extraction by using the alternatives, as shown in Fig.~\ref{fig:distance}.
We see that the Global position strategy shows the lowest helpfulness consistently, compared to the relative position methods.
This finding suggests that relative distance may be more helpful in modeling the conversation discourse.
Moreover, the RoPE gives the best usefulness, especially under inter-utterance cases.
Intuitively, such dynamic position information offers more flexible bridging knowledge for easing the long-range dependence issue of term pairing where the entities are separated in different utterances in distance.

\vspace{-1mm}
\section{What To Do Next?}

\vspace{-1mm}
In this work, we propose an initial method to solve the DiaASQ task.
Although achieving stronger performances than baselines, it could be further benefited from many angles.
To facilitate the follow-up research in this direction, we try to shed light on several potential future works.

\vspace{4pt}
\noindent$\blacktriangleright$ \textbf{Making Better Use of The Dialogue Discourse Structure Information.}
The core challenge of the DiaASQ task lies in handling conversation contexts.
Compared to the typical case of single sentences, the dialogue utterances are syntactically disjoint.
Thus, it is critical to carefully model the dialogue discourse structure information~\cite{FeiDiaREIJCAI22}, so as to better capture the dialogue semantics, for better recognition of the cross-utterance quadruples.
Although we leverage the dialogue relative distance information (RoPE) in this work, without treating the dialogue utterances as a whole, our method may still lose some important discourse information.
As seen in Fig.~\ref{fig:cross}, our model's performance on the super cross-utterance quads is still far from satisfaction, i.e., zero F1 score on the cross-$>$3-utterance case. 
Intuitively, constructing an explicit conversational discourse structure (i.e., the tree or graph structure) for the task is promising.

\vspace{4pt}
\noindent$\blacktriangleright$ \textbf{Enhancing Coreference Resolution.}
In the conversation scenario, the speaker and target coreference is one of the biggest issues.
In the DiaASQ task, the bundled sentiment elements (e.g., target, aspect, and opinion term) of one quad may be yielded by different users or maybe one individual.
Besides, the sentiment terms may be coreferred by pronouns, for example, `\emph{the screen quality of it}' where `\emph{it}' refers to the target term `\emph{Xiaomi 6}' mentioned in the previous context.
Without correctly understanding the coreference, it is problematic for a system to precisely capture the context semantics, and thus leads to a wrong pairing between sentiment elements and unexplainable predictions.

\vspace{4pt}
\noindent$\blacktriangleright$ \textbf{Extracting Overlapped Quadruple.}
It is common in our DiaASQ dataset that one sentiment term of one quad overlaps with other terms of another quad.
For example, different electronic devices (targets) may have the same aspects, e.g., battery life, screen or size, etc.
A sound DiaASQ system should also well solve the quadruple overlap issue.
We note that the overlapped quads can essentially share certain structural information, and thus it is favorable to use such shared knowledge effectively.

\vspace{4pt}
\noindent$\blacktriangleright$ \textbf{Transferring Well-learned Sentiment Knowledge from Existing System.}
The sentiment analysis community has developed a great amount of powerful ABSA systems well-trained on the large-scale free texts or existing sentiment corpora~\cite{XuSYL20,TianGXLHWWW20,LiZZZW21}.
Since this work still inherits the basic spirit of ABSA, it is naturally a promising idea to transfer the existing well-trained sentiment-enriched ABSA model for enhancing the understanding of the DiaASQ task.

\vspace{4pt}
\noindent$\blacktriangleright$ \textbf{Multi-/Cross-lingual Dialogue ABSA.}
One of the key challenges for more accurate multi-/cross-lingual ABSA is the missing of parallel annotations in different languages, i.e., causing troubles for label alignments~\cite{feng-wan-2019-learning,fei-li-2020-cross,zhang-etal-2021-cross}.
As we annotate the DiaASQ dataset in two languages (i.e., Chinese and English) with parallel sentences, this paves the way for the research of more effective multi-lingual or cross-lingual dialogue-level ABSA.

\vspace{-2mm}
\section{Conclusion} 

\vspace{-1mm}
This work introduces a new task of conversational aspect-based sentiment quadruple analysis, namely DiaASQ, which aims to detect the sentiment quadruple of \emph{target-aspect-opinion-sentiment} structure in the conversation texts.
DiaASQ bridges the gap between conversational opinion mining and fine-grained sentiment analysis.
We manually construct a large-scale, high-quality dataset with Chinese and English versions for the task, with 1,000 dialogue snippets, including 7,452 utterances.
We then benchmark the DiaASQ task with an end-to-end neural model, which effectively models the dialogue utterance interactions.
Experiments demonstrate the advantages of our method in effectively learning the dialogue-specific features for better cross-utterance sentiment quadruple extraction.

\section*{Acknowledgment}
This work is supported by 
the National Key Research and Development Program of China (No. 2022YFB3103602, No. 2017YFC1200500),
the National Natural Science Foundation of China (No. 62176187),
the Research Foundation of Ministry of Education of China (No. 18JZD015),
China Scholarship Council (CSC),
and Sea-NExT Joint Lab.

\section*{Limitations}
Our paper has the following potential limitations.
First, our current DiaASQ dataset is limited to only the domain of digital devices.
We plan to further extend the DiaASQ texts to other domains, e.g., foods/restaurants, hotel/trips, etc.
Secondly, our proposed model may be limited to insufficient modeling of the dialogue-level discourse structure information, which would somehow prevent us from obtaining further task improvements.
Third, in DiaASQ task, it is more difficult to recognize the opinion terms, compared to the extraction of target and aspect terms.
This may largely deteriorate the overall performance due to the fact that opinion expressions are much more flexible and sometimes are subject to satirical expression.

\section*{Ethical Considerations}
Here we discuss the primary ethical considerations of the DiaASQ dataset.

\vspace{-4pt}
\paragraph{Intellectual Property Protection.}
Our dataset is collected from the open Chinese social media platform via the officially open API.\footnote{\url{https://open.weibo.com/wiki/API}}
Permissions are granted to copy, distribute and modify the contents under the terms of Weibo API distribution.

\vspace{-4pt}
\paragraph{Privacy Claim.}
The user-specific information in the data is anonymized during preprocessing, and no personal information of the user or customer is included.
The data collection procedure is designed for factual knowledge acquisition and does not involve privacy issues.

\vspace{-4pt}
\paragraph{Annotator Information and Compensation.}
The crowd-sourcing annotators are the senior postgraduate students who are trained before annotating.
We estimated that a skillful annotator needs 3 to 5 minutes to finish an annotation for each dialogue utterance.
Therefore, we paid annotators 1 yuan (\$0.15) for each utterance. 
The salaries for linguistic and computer science experts are determined by the average time they devote.

\bibliography{anthology}

\begin{thebibliography}{61}
\expandafter\ifx\csname natexlab\endcsname\relax\def\natexlab#1{#1}\fi

\bibitem[{Barnes et~al.(2021)Barnes, Kurtz, Oepen, {\O}vrelid, and
  Velldal}]{barnes-etal-2021-structured}
Jeremy Barnes, Robin Kurtz, Stephan Oepen, Lilja {\O}vrelid, and Erik Velldal.
  2021.
\newblock \href {https://doi.org/10.18653/v1/2021.acl-long.263} {Structured
  sentiment analysis as dependency graph parsing}.
\newblock In \emph{Proc. of ACL}, pages 3387--3402.

\bibitem[{Cai et~al.(2021)Cai, Xia, and Yu}]{CaiXY20}
Hongjie Cai, Rui Xia, and Jianfei Yu. 2021.
\newblock \href {https://doi.org/10.18653/v1/2021.acl-long.29}
  {Aspect-category-opinion-sentiment quadruple extraction with implicit aspects
  and opinions}.
\newblock In \emph{Proc. of ACL}, pages 340--350.

\bibitem[{Cambria(2016)}]{Cambria16}
Erik Cambria. 2016.
\newblock \href {https://doi.org/10.1109/MCI.2019.2901082} {Affective computing
  and sentiment analysis}.
\newblock \emph{{IEEE} Intell. Syst.}, 31(2):102--107.

\bibitem[{Chen et~al.(2022{\natexlab{a}})Chen, Zhai, Feng, Li, and
  Wang}]{ChenZFLW22}
Hao Chen, Zepeng Zhai, Fangxiang Feng, Ruifan Li, and Xiaojie Wang.
  2022{\natexlab{a}}.
\newblock \href {https://doi.org/10.18653/v1/2022.acl-long.212} {Enhanced
  multi-channel graph convolutional network for aspect sentiment triplet
  extraction}.
\newblock In \emph{Proc. of ACL}, pages 2974--2985.

\bibitem[{Chen et~al.(2020)Chen, Liu, Wang, Zhang, and Chi}]{ChenLWZC20}
Shaowei Chen, Jie Liu, Yu~Wang, Wenzheng Zhang, and Ziming Chi. 2020.
\newblock \href {https://doi.org/10.18653/v1/2020.acl-main.582} {Synchronous
  double-channel recurrent network for aspect-opinion pair extraction}.
\newblock In \emph{Proc. of ACL}, pages 6515--6524.

\bibitem[{Chen et~al.(2022{\natexlab{b}})Chen, Keming, Sun, and
  Zhang}]{chen-2022-span}
Yuqi Chen, Chen Keming, Xian Sun, and Zequn Zhang. 2022{\natexlab{b}}.
\newblock \href {https://aclanthology.org/2022.emnlp-main.289} {A span-level
  bidirectional network for aspect sentiment triplet extraction}.
\newblock In \emph{Proc. of EMNLP}, pages 4300--4309.

\bibitem[{Chen et~al.(2021)Chen, Huang, Liu, Shi, and Jin}]{ChenHLSJ21}
Zhexue Chen, Hong Huang, Bang Liu, Xuanhua Shi, and Hai Jin. 2021.
\newblock \href {https://doi.org/10.18653/v1/2021.findings-acl.128} {Semantic
  and syntactic enhanced aspect sentiment triplet extraction}.
\newblock In \emph{Proc. of ACL Findings}, pages 1474--1483.

\bibitem[{Cui et~al.(2021)Cui, Che, Liu, Qin, and Yang}]{CuiCLQY21}
Yiming Cui, Wanxiang Che, Ting Liu, Bing Qin, and Ziqing Yang. 2021.
\newblock \href {https://doi.org/10.1109/TASLP.2021.3124365} {Pre-training with
  whole word masking for chinese {BERT}}.
\newblock \emph{{IEEE} {ACM} Trans. Audio Speech Lang. Process.}, pages
  3504--3514.

\bibitem[{Dai et~al.(2020)Dai, Peng, Chen, and Ding}]{dai-etal-2020-multi}
Zehui Dai, Cheng Peng, Huajie Chen, and Yadong Ding. 2020.
\newblock \href {https://doi.org/10.18653/v1/2020.emnlp-main.565} {A multi-task
  incremental learning framework with category name embedding for
  aspect-category sentiment analysis}.
\newblock In \emph{Proc. of EMNLP}, pages 6955--6965.

\bibitem[{Devlin et~al.(2019)Devlin, Chang, Lee, and Toutanova}]{DevlinCLT19}
Jacob Devlin, Ming{-}Wei Chang, Kenton Lee, and Kristina Toutanova. 2019.
\newblock {BERT:} pre-training of deep bidirectional transformers for language
  understanding.
\newblock In \emph{Proc. of NAACL}, pages 4171--4186.

\bibitem[{Dou and Neubig(2021)}]{dou-neubig-2021-word}
Zi-Yi Dou and Graham Neubig. 2021.
\newblock \href {https://doi.org/10.18653/v1/2021.eacl-main.181} {Word
  alignment by fine-tuning embeddings on parallel corpora}.
\newblock In \emph{Proc. of EMNLP}, pages 2112--2128.

\bibitem[{Eberts and Ulges(2020)}]{EbertsU20}
Markus Eberts and Adrian Ulges. 2020.
\newblock \href {https://doi.org/10.3233/FAIA200321} {Span-based joint entity
  and relation extraction with transformer pre-training}.
\newblock In \emph{Proc. of ECAI}, pages 2006--2013.

\bibitem[{Fan et~al.(2018)Fan, Feng, and Zhao}]{FanFZ18}
Feifan Fan, Yansong Feng, and Dongyan Zhao. 2018.
\newblock \href {https://doi.org/10.18653/v1/d18-1380} {Multi-grained attention
  network for aspect-level sentiment classification}.
\newblock In \emph{Proc. of EMNLP}, pages 3433--3442.

\bibitem[{Fan et~al.(2019)Fan, Wu, Dai, Huang, and Chen}]{fan-etal-2019-target}
Zhifang Fan, Zhen Wu, Xin{-}Yu Dai, Shujian Huang, and Jiajun Chen. 2019.
\newblock \href {https://doi.org/10.18653/v1/n19-1259} {Target-oriented opinion
  words extraction with target-fused neural sequence labeling}.
\newblock In \emph{Proc. of NAACL}, pages 2509--2518.

\bibitem[{Fei et~al.(2022{\natexlab{a}})Fei, Li, Li, Wu, Li, and
  Ji}]{Feiijcai22UABSA}
Hao Fei, Fei Li, Chenliang Li, Shengqiong Wu, Jingye Li, and Donghong Ji.
  2022{\natexlab{a}}.
\newblock \href {https://doi.org/10.24963/ijcai.2022/572} {Inheriting the
  wisdom of predecessors: A multiplex cascade framework for unified
  aspect-based sentiment analysis}.
\newblock In \emph{Proc. of IJCAI}, pages 4096--4103.

\bibitem[{Fei et~al.(2022{\natexlab{b}})Fei, Li, Wu, Li, Ji, and
  Li}]{FeiDiaREIJCAI22}
Hao Fei, Jingye Li, Shengqiong Wu, Chenliang Li, Donghong Ji, and Fei Li.
  2022{\natexlab{b}}.
\newblock Global inference with explicit syntactic and discourse structures for
  dialogue-level relation extraction.
\newblock In \emph{Proc. of IJCAI}, pages 4082--4088.

\bibitem[{Fei et~al.(2020)Fei, Zhang, and Ji}]{fei-etal-2020-cross}
Hao Fei, Meishan Zhang, and Donghong Ji. 2020.
\newblock \href {https://doi.org/10.18653/v1/2020.acl-main.627} {Cross-lingual
  semantic role labeling with high-quality translated training corpus}.
\newblock In \emph{Proc. of ACL}, pages 7014--7026.

\bibitem[{Fei and Li(2020)}]{fei-li-2020-cross}
Hongliang Fei and Ping Li. 2020.
\newblock \href {https://doi.org/10.18653/v1/2020.acl-main.510} {Cross-lingual
  unsupervised sentiment classification with multi-view transfer learning}.
\newblock In \emph{Proc. of ACL}, pages 5759--5771.

\bibitem[{Feng and Wan(2019)}]{feng-wan-2019-learning}
Yanlin Feng and Xiaojun Wan. 2019.
\newblock \href {https://doi.org/10.18653/v1/n19-1040} {Learning bilingual
  sentiment-specific word embeddings without cross-lingual supervision}.
\newblock In \emph{Proc. of NAACL}, pages 420--429.

\bibitem[{Hu et~al.(2021)Hu, Wei, and Huai}]{hu-etal-2021-dialoguecrn}
Dou Hu, Lingwei Wei, and Xiaoyong Huai. 2021.
\newblock \href {https://doi.org/10.18653/v1/2021.acl-long.547} {Dialoguecrn:
  Contextual reasoning networks for emotion recognition in conversations}.
\newblock In \emph{Proc. of ACL}, pages 7042--7052.

\bibitem[{Jiang et~al.(2019)Jiang, Chen, Xu, Ao, and
  Yang}]{jiang-etal-2019-challenge}
Qingnan Jiang, Lei Chen, Ruifeng Xu, Xiang Ao, and Min Yang. 2019.
\newblock \href {https://doi.org/10.18653/v1/D19-1654} {A challenge dataset and
  effective models for aspect-based sentiment analysis}.
\newblock In \emph{Proc. of EMNLP}, pages 6279--6284.

\bibitem[{Li et~al.(2020)Li, Fei, and Ji}]{li-etal-2020-modeling}
Jingye Li, Hao Fei, and Donghong Ji. 2020.
\newblock \href {https://doi.org/10.18653/v1/2020.coling-main.53} {Modeling
  local contexts for joint dialogue act recognition and sentiment
  classification with bi-channel dynamic convolutions}.
\newblock In \emph{Proc. of COLING}, pages 616--626.

\bibitem[{Li et~al.(2019)Li, Bing, Li, and Lam}]{li2021unified}
Xin Li, Lidong Bing, Piji Li, and Wai Lam. 2019.
\newblock \href {https://doi.org/10.1609/aaai.v33i01.33016714} {A unified model
  for opinion target extraction and target sentiment prediction}.
\newblock In \emph{Proc. of AAAI}, pages 6714--6721.

\bibitem[{Li et~al.(2018)Li, Bing, Li, Lam, and Yang}]{LiBLLY18}
Xin Li, Lidong Bing, Piji Li, Wai Lam, and Zhimou Yang. 2018.
\newblock \href {https://doi.org/10.24963/ijcai.2018/583} {Aspect term
  extraction with history attention and selective transformation}.
\newblock In \emph{Proc. of IJCAI}, pages 4194--4200.

\bibitem[{Li et~al.(2022)Li, Tang, Zhao, and Zhu}]{li-etal-2022-emocaps}
Zaijing Li, Fengxiao Tang, Ming Zhao, and Yusen Zhu. 2022.
\newblock \href {https://doi.org/10.18653/v1/2022.findings-acl.126} {Emocaps:
  Emotion capsule based model for conversational emotion recognition}.
\newblock In \emph{Proc. of ACL Findings}, pages 1610--1618.

\bibitem[{Li et~al.(2021)Li, Zou, Zhang, Zhang, and Wei}]{LiZZZW21}
Zhengyan Li, Yicheng Zou, Chong Zhang, Qi~Zhang, and Zhongyu Wei. 2021.
\newblock \href {https://doi.org/10.18653/v1/2021.emnlp-main.22} {Learning
  implicit sentiment in aspect-based sentiment analysis with supervised
  contrastive pre-training}.
\newblock In \emph{Proc. of EMNLP}, pages 246--256.

\bibitem[{Liao et~al.(2021)Liao, Long, Zhang, Huang, and Chua}]{LiaoLZHC21}
Lizi Liao, Le~Hong Long, Zheng Zhang, Minlie Huang, and Tat{-}Seng Chua. 2021.
\newblock \href {https://doi.org/10.1145/3477495.3532063} {Mmconv: An
  environment for multimodal conversational search across multiple domains}.
\newblock In \emph{Proc. of SIGIR}, pages 675--684.

\bibitem[{Liao et~al.(2022)Liao, Takanobu, Ma, Yang, Huang, and
  Chua}]{LiaoTMYHC22}
Lizi Liao, Ryuichi Takanobu, Yunshan Ma, Xun Yang, Minlie Huang, and Tat{-}Seng
  Chua. 2022.
\newblock \href {https://doi.org/10.1109/TKDE.2020.3008563} {Topic-guided
  conversational recommender in multiple domains}.
\newblock \emph{{IEEE} Trans. Knowl. Data Eng.}, 34(5):2485--2496.

\bibitem[{Liu et~al.(2021{\natexlab{a}})Liu, Zheng, Demasi, Sabour, Li, Yu,
  Jiang, and Huang}]{liu-etal-2021-towards}
Siyang Liu, Chujie Zheng, Orianna Demasi, Sahand Sabour, Yu~Li, Zhou Yu, Yong
  Jiang, and Minlie Huang. 2021{\natexlab{a}}.
\newblock \href {https://doi.org/10.18653/v1/2021.acl-long.269} {Towards
  emotional support dialog systems}.
\newblock In \emph{Proc. of ACL}, pages 3469--3483.

\bibitem[{Liu et~al.(2019)Liu, Ott, Goyal, Du, Joshi, Chen, Levy, Lewis,
  Zettlemoyer, and Stoyanov}]{roberta-liu}
Yinhan Liu, Myle Ott, Naman Goyal, Jingfei Du, Mandar Joshi, Danqi Chen, Omer
  Levy, Mike Lewis, Luke Zettlemoyer, and Veselin Stoyanov. 2019.
\newblock \href {http://arxiv.org/abs/1907.11692} {Roberta: {A} robustly
  optimized {BERT} pretraining approach}.
\newblock \emph{CoRR}.

\bibitem[{Liu et~al.(2021{\natexlab{b}})Liu, Xia, and
  Yu}]{liu-etal-2021-comparative}
Ziheng Liu, Rui Xia, and Jianfei Yu. 2021{\natexlab{b}}.
\newblock \href {https://doi.org/10.18653/v1/2021.emnlp-main.322} {Comparative
  opinion quintuple extraction from product reviews}.
\newblock In \emph{Proc. of EMNLP}, pages 3955--3965.

\bibitem[{McDonald et~al.(2007)McDonald, Hannan, Neylon, Wells, and
  Reynar}]{mcdonald-etal-2007-structured}
Ryan McDonald, Kerry Hannan, Tyler Neylon, Mike Wells, and Jeff Reynar. 2007.
\newblock \href {https://aclanthology.org/P07-1055/} {Structured models for
  fine-to-coarse sentiment analysis}.
\newblock In \emph{Proc. of ACL}, pages 432--439.

\bibitem[{Mukherjee et~al.(2021)Mukherjee, Nayak, Butala, Bhattacharya, and
  Goyal}]{mu-etal-21-pte}
Rajdeep Mukherjee, Tapas Nayak, Yash Butala, Sourangshu Bhattacharya, and Pawan
  Goyal. 2021.
\newblock \href {https://doi.org/10.18653/v1/2021.emnlp-main.731} {{PASTE:} {A}
  tagging-free decoding framework using pointer networks for aspect sentiment
  triplet extraction}.
\newblock In \emph{Proc. of EMNLP}, pages 9279--9291.

\bibitem[{Ni et~al.(2022)Ni, Young, Pandelea, Xue, and Cambria}]{ni2022recent}
Jinjie Ni, Tom Young, Vlad Pandelea, Fuzhao Xue, and Erik Cambria. 2022.
\newblock \href {https://doi.org/https://doi.org/10.1007/s10462-022-10248-8}
  {Recent advances in deep learning based dialogue systems: A systematic
  survey}.
\newblock \emph{Artificial intelligence review}, pages 1--101.

\bibitem[{Pang and Lee(2007)}]{PangL07}
Bo~Pang and Lillian Lee. 2007.
\newblock \href {https://doi.org/10.1561/1500000011} {Opinion mining and
  sentiment analysis}.
\newblock \emph{Foundations and Trends in Information Retrieval},
  2(1-2):1--135.

\bibitem[{Peng et~al.(2020)Peng, Xu, Bing, Huang, Lu, and Si}]{PengXBHLS20}
Haiyun Peng, Lu~Xu, Lidong Bing, Fei Huang, Wei Lu, and Luo Si. 2020.
\newblock \href {https://ojs.aaai.org/index.php/AAAI/article/view/6383}
  {Knowing what, how and why: {A} near complete solution for aspect-based
  sentiment analysis}.
\newblock In \emph{Proc. of AAAI}, pages 8600--8607.

\bibitem[{Pontiki et~al.(2016)Pontiki, Galanis, Papageorgiou, Androutsopoulos,
  Manandhar, Al{-}Smadi, Al{-}Ayyoub, Zhao, Qin, Clercq, Hoste, Apidianaki,
  Tannier, Loukachevitch, Kotelnikov, Bel, Zafra, and
  Eryigit}]{PontikiGPAMAAZQ16}
Maria Pontiki, Dimitris Galanis, Haris Papageorgiou, Ion Androutsopoulos,
  Suresh Manandhar, Mohammad Al{-}Smadi, Mahmoud Al{-}Ayyoub, Yanyan Zhao, Bing
  Qin, Orph{\'{e}}e~De Clercq, V{\'{e}}ronique Hoste, Marianna Apidianaki,
  Xavier Tannier, Natalia~V. Loukachevitch, Evgeniy~V. Kotelnikov, N{\'{u}}ria
  Bel, Salud Mar{\'{\i}}a~Jim{\'{e}}nez Zafra, and G{\"{u}}lsen Eryigit. 2016.
\newblock \href {https://doi.org/10.18653/v1/s16-1002} {Semeval-2016 task 5:
  Aspect based sentiment analysis}.
\newblock In \emph{Proc. of SemEval}, pages 19--30.

\bibitem[{Pontiki et~al.(2015)Pontiki, Galanis, Papageorgiou, Manandhar, and
  Androutsopoulos}]{PontikiGPMA15}
Maria Pontiki, Dimitris Galanis, Haris Papageorgiou, Suresh Manandhar, and Ion
  Androutsopoulos. 2015.
\newblock \href {https://doi.org/10.18653/v1/s15-2082} {Semeval-2015 task 12:
  Aspect based sentiment analysis}.
\newblock In \emph{Proc. of SemEval}, pages 486--495.

\bibitem[{Pontiki et~al.(2014)Pontiki, Galanis, Pavlopoulos, Papageorgiou,
  Androutsopoulos, and Manandhar}]{PontikiGPPAM14}
Maria Pontiki, Dimitris Galanis, John Pavlopoulos, Harris Papageorgiou, Ion
  Androutsopoulos, and Suresh Manandhar. 2014.
\newblock \href {https://doi.org/10.3115/v1/s14-2004} {Semeval-2014 task 4:
  Aspect based sentiment analysis}.
\newblock In \emph{Proc. of SemEval}, pages 27--35.

\bibitem[{Ren et~al.(2016)Ren, Zhang, Zhang, and Ji}]{RenZZJ16}
Yafeng Ren, Yue Zhang, Meishan Zhang, and Donghong Ji. 2016.
\newblock \href
  {http://www.aaai.org/ocs/index.php/AAAI/AAAI16/paper/view/11922}
  {Context-sensitive twitter sentiment classification using neural network}.
\newblock In \emph{Proc. of AAAI}, pages 215--221.

\bibitem[{Shen et~al.(2021)Shen, Chen, Quan, and Xie}]{ShenCQX21}
Weizhou Shen, Junqing Chen, Xiaojun Quan, and Zhixian Xie. 2021.
\newblock \href {https://ojs.aaai.org/index.php/AAAI/article/view/17625}
  {Dialogxl: All-in-one xlnet for multi-party conversation emotion
  recognition}.
\newblock In \emph{Proc. of AAAI}, pages 13789--13797.

\bibitem[{Shi et~al.(2022)Shi, Li, Li, Fei, and Ji}]{ShiLL0J22}
Wenxuan Shi, Fei Li, Jingye Li, Hao Fei, and Donghong Ji. 2022.
\newblock \href {https://doi.org/10.18653/v1/2022.acl-long.291} {Effective
  token graph modeling using a novel labeling strategy for structured sentiment
  analysis}.
\newblock In \emph{Proc. of ACL}, pages 4232--4241.

\bibitem[{Song et~al.(2022)Song, Xin, Lai, Wang, Su, and Xu}]{song-casa-2022}
Linfeng Song, Chunlei Xin, Shaopeng Lai, Ante Wang, Jinsong Su, and Kun Xu.
  2022.
\newblock \href {https://doi.org/10.1613/jair.1.12802} {{CASA:} conversational
  aspect sentiment analysis for dialogue understanding}.
\newblock \emph{J. Artif. Intell. Res.}, 73:511--533.

\bibitem[{Su et~al.(2021)Su, Lu, Pan, Wen, and Liu}]{abs-2104-09864}
Jianlin Su, Yu~Lu, Shengfeng Pan, Bo~Wen, and Yunfeng Liu. 2021.
\newblock \href {https://arxiv.org/abs/2104.09864} {Roformer: Enhanced
  transformer with rotary position embedding}.
\newblock \emph{CoRR}.

\bibitem[{Tang et~al.(2016)Tang, Qin, Feng, and Liu}]{tang-etal-2016-effective}
Duyu Tang, Bing Qin, Xiaocheng Feng, and Ting Liu. 2016.
\newblock \href {https://aclanthology.org/C16-1311/} {Effective lstms for
  target-dependent sentiment classification}.
\newblock In \emph{Proc. of COLING}, pages 3298--3307.

\bibitem[{Tian et~al.(2020)Tian, Gao, Xiao, Liu, He, Wu, Wang, and
  Wu}]{TianGXLHWWW20}
Hao Tian, Can Gao, Xinyan Xiao, Hao Liu, Bolei He, Hua Wu, Haifeng Wang, and
  Feng Wu. 2020.
\newblock \href {https://doi.org/10.18653/v1/2020.acl-main.374} {{SKEP:}
  sentiment knowledge enhanced pre-training for sentiment analysis}.
\newblock In \emph{Proc. of ACL}, pages 4067--4076.

\bibitem[{Vaswani et~al.(2017)Vaswani, Shazeer, Parmar, Uszkoreit, Jones,
  Gomez, Kaiser, and Polosukhin}]{VaswaniSPUJGKP17}
Ashish Vaswani, Noam Shazeer, Niki Parmar, Jakob Uszkoreit, Llion Jones,
  Aidan~N. Gomez, Lukasz Kaiser, and Illia Polosukhin. 2017.
\newblock \href
  {https://proceedings.neurips.cc/paper/2017/hash/3f5ee243547dee91fbd053c1c4a845aa-Abstract.html}
  {Attention is all you need}.
\newblock In \emph{Proc. of NeurIPS}, pages 5998--6008.

\bibitem[{Wang et~al.(2019)Wang, Sun, Li, Liu, Si, Zhang, and
  Zhou}]{wang-2019-aspect}
Jingjing Wang, Changlong Sun, Shoushan Li, Xiaozhong Liu, Luo Si, Min Zhang,
  and Guodong Zhou. 2019.
\newblock \href {https://doi.org/10.18653/v1/P19-1345} {Aspect sentiment
  classification towards question-answering with reinforced bidirectional
  attention network}.
\newblock In \emph{Proc. of ACL}, pages 3548--3557.

\bibitem[{Wu et~al.(2022)Wu, Fei, Li, Zhang, Liu, Teng, and Ji}]{Wu0LZLTJ22}
Shengqiong Wu, Hao Fei, Fei Li, Meishan Zhang, Yijiang Liu, Chong Teng, and
  Donghong Ji. 2022.
\newblock \href {https://ojs.aaai.org/index.php/AAAI/article/view/21404}
  {Mastering the explicit opinion-role interaction: Syntax-aided neural
  transition system for unified opinion role labeling}.
\newblock In \emph{Proc. of AAAI}, pages 11513--11521.

\bibitem[{Wu et~al.(2021)Wu, Fei, Ren, Ji, and Li}]{Wu0RJL21}
Shengqiong Wu, Hao Fei, Yafeng Ren, Donghong Ji, and Jingye Li. 2021.
\newblock \href {https://doi.org/10.24963/ijcai.2021/545} {Learn from syntax:
  Improving pair-wise aspect and opinion terms extraction with rich syntactic
  knowledge}.
\newblock In \emph{Proc. of IJCAI}, pages 3957--3963.

\bibitem[{Wu et~al.(2020)Wu, Ying, Zhao, Fan, Dai, and Xia}]{wu-etal-2020-grid}
Zhen Wu, Chengcan Ying, Fei Zhao, Zhifang Fan, Xinyu Dai, and Rui Xia. 2020.
\newblock \href {https://arxiv.org/abs/2010.04640} {Grid tagging scheme for
  aspect-oriented fine-grained opinion extraction}.
\newblock In \emph{Proc. of EMNLP Findings}, pages 2576--2585.

\bibitem[{Xu et~al.(2020)Xu, Shu, Yu, and Liu}]{XuSYL20}
Hu~Xu, Lei Shu, Philip~S. Yu, and Bing Liu. 2020.
\newblock \href {https://doi.org/10.18653/v1/2020.coling-main.21}
  {Understanding pre-trained {BERT} for aspect-based sentiment analysis}.
\newblock In \emph{Proc. of COLING}, pages 244--250.

\bibitem[{Xu et~al.(2021)Xu, Chia, and Bing}]{XuCB20}
Lu~Xu, Yew~Ken Chia, and Lidong Bing. 2021.
\newblock \href {https://doi.org/10.18653/v1/2021.acl-long.367} {Learning
  span-level interactions for aspect sentiment triplet extraction}.
\newblock In \emph{Proc. of ACL}, pages 4755--4766.

\bibitem[{Xue et~al.(2021)Xue, Constant, Roberts, Kale, Al{-}Rfou, Siddhant,
  Barua, and Raffel}]{XueCRKASBR21}
Linting Xue, Noah Constant, Adam Roberts, Mihir Kale, Rami Al{-}Rfou, Aditya
  Siddhant, Aditya Barua, and Colin Raffel. 2021.
\newblock \href {https://doi.org/10.18653/v1/2021.naacl-main.41} {mt5: {A}
  massively multilingual pre-trained text-to-text transformer}.
\newblock In \emph{Proc. of NAACL}, pages 483--498.

\bibitem[{Zhang et~al.(2021{\natexlab{a}})Zhang, Deng, Li, Yuan, Bing, and
  Lam}]{ZhangD0YBL21}
Wenxuan Zhang, Yang Deng, Xin Li, Yifei Yuan, Lidong Bing, and Wai Lam.
  2021{\natexlab{a}}.
\newblock \href {https://doi.org/10.18653/v1/2021.emnlp-main.726} {Aspect
  sentiment quad prediction as paraphrase generation}.
\newblock In \emph{Proc. of EMNLP}, pages 9209--9219.

\bibitem[{Zhang et~al.(2021{\natexlab{b}})Zhang, He, Peng, Bing, and
  Lam}]{zhang-etal-2021-cross}
Wenxuan Zhang, Ruidan He, Haiyun Peng, Lidong Bing, and Wai Lam.
  2021{\natexlab{b}}.
\newblock \href {https://doi.org/10.18653/v1/2021.emnlp-main.727}
  {Cross-lingual aspect-based sentiment analysis with aspect term
  code-switching}.
\newblock In \emph{Proc. of EMNLP}, pages 9220--9230.

\bibitem[{Zhang et~al.(2021{\natexlab{c}})Zhang, Li, Deng, Bing, and
  Lam}]{Zhang0DBL20}
Wenxuan Zhang, Xin Li, Yang Deng, Lidong Bing, and Wai Lam. 2021{\natexlab{c}}.
\newblock \href {https://doi.org/10.18653/v1/2021.acl-short.64} {Towards
  generative aspect-based sentiment analysis}.
\newblock In \emph{Proc. of ACL}, pages 504--510.

\bibitem[{Zhao et~al.(2020)Zhao, Huang, Zhang, Lu, and
  Xue}]{zhao-etal-2020-spanmlt}
He~Zhao, Longtao Huang, Rong Zhang, Quan Lu, and Hui Xue. 2020.
\newblock \href {https://doi.org/10.18653/v1/2020.acl-main.296} {Spanmlt: {A}
  span-based multi-task learning framework for pair-wise aspect and opinion
  terms extraction}.
\newblock In \emph{Proc. of ACL}, pages 3239--3248.

\bibitem[{Zhao et~al.(2022)Zhao, Zhang, Hu, Liu, Jin, Wang, and
  Li}]{ZhaoZ0LJW022}
Jinming Zhao, Tenggan Zhang, Jingwen Hu, Yuchen Liu, Qin Jin, Xinchao Wang, and
  Haizhou Li. 2022.
\newblock \href {https://doi.org/10.18653/v1/2022.acl-long.391} {{M3ED:}
  multi-modal multi-scene multi-label emotional dialogue database}.
\newblock In \emph{Proc. of ACL}, pages 5699--5710.

\bibitem[{Zhen et~al.(2021)Zhen, Wang, Fu, Lv, and
  Zhang}]{zhen-etal-2021-chinese}
Ranran Zhen, Rui Wang, Guohong Fu, Chengguo Lv, and Meishan Zhang. 2021.
\newblock \href {https://doi.org/10.18653/v1/2021.emnlp-main.796} {{C}hinese
  opinion role labeling with corpus translation: A pivot study}.
\newblock In \emph{Proc. of EMNLP}, pages 10139--10149.

\bibitem[{Zhou et~al.(2021)Zhou, Liao, Gao, Jie, and Lu}]{zhu-tal-01-clser}
Yuxiang Zhou, Lejian Liao, Yang Gao, Zhanming Jie, and Wei Lu. 2021.
\newblock \href {https://doi.org/10.18653/v1/2021.emnlp-main.317} {To be
  closer: Learning to link up aspects with opinions}.
\newblock In \emph{Proc. of EMNLP}, pages 3899--3909.

\end{thebibliography}
\bibliographystyle{acl_natbib}

\newpage

\appendix

\clearpage

\appendix

\section{Model and Setup Specification}

\begin{algorithm}[h]
	\caption{Calculating global indices of tokens in two threads} 
	\label{alg.rela} 
	\begin{algorithmic}
	    \REQUIRE $P_t$; Two thread $T_i,T_j$, where $i,j$ are thread id.
	    \IF {$i * j$ \text{==} $0$ \OR $i \text{==} j$}
	    \STATE $P^{ij}_t=P_t(t\in {T_i,T_j})$
	    \ELSIF{i < j}
	    \STATE $P^{ij}_t=-P_t(t\in {T_i})$
	    \STATE $P^{ij}_t=P_t(t\in {T_j})$
	    \ELSE
	    \STATE $P^{ij}_t=P_t(t\in {T_i})$
	    \STATE $P^{ij}_t=-P_t(t\in {T_j})$
	   \ENDIF
		
	\end{algorithmic} 
\end{algorithm}

\paragraph{Distance Encoding Details.}
\label{app:rope_detail}

\begin{figure}[!ht]
  \centering
  \includegraphics[width=0.74\columnwidth]{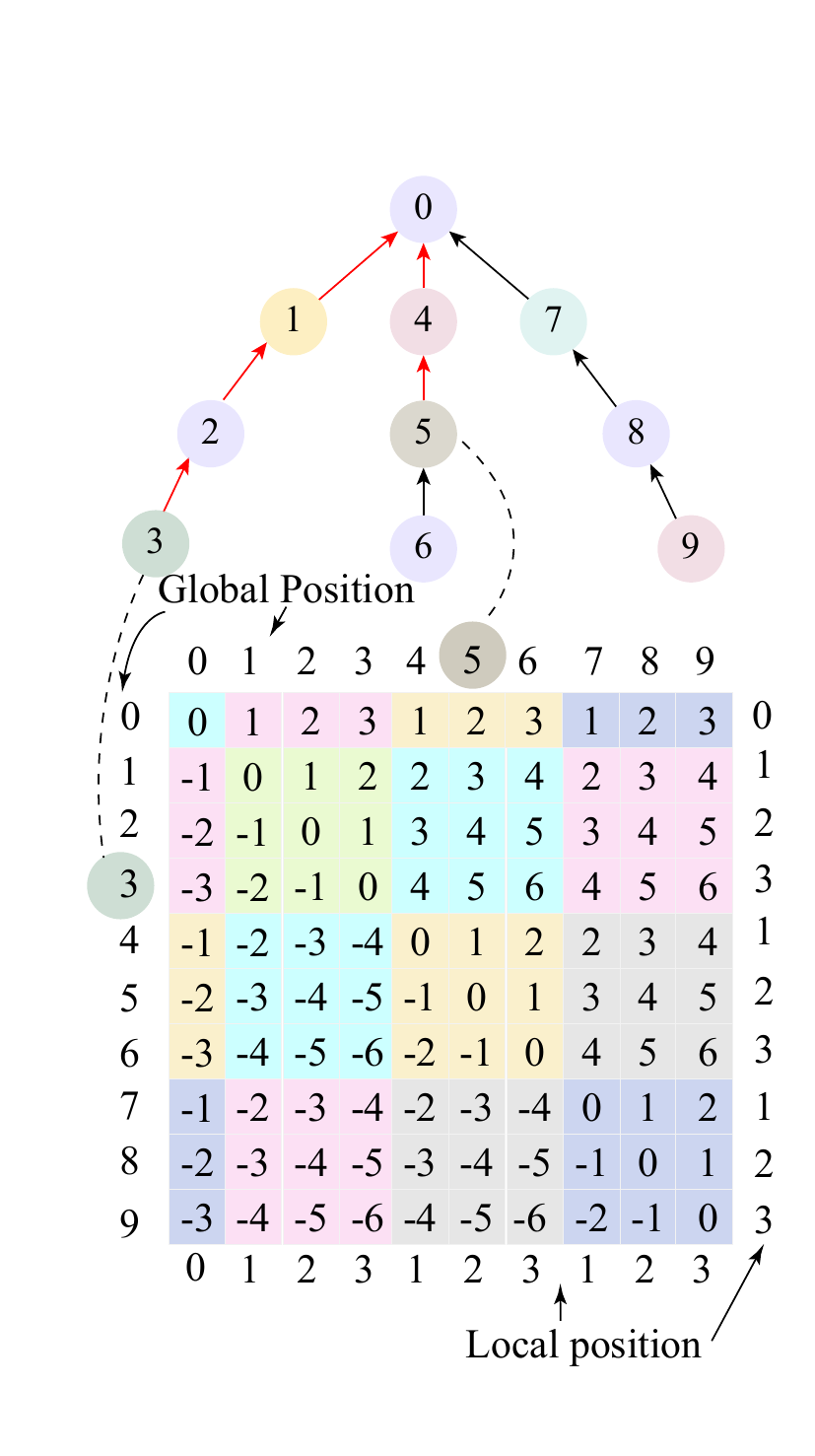}
  \caption{
Relative token distance calculation over the dialogue tree structure.
To simplify the problem, we assume that one utterance has only one token.
}
  \label{fig:distance_matrix}
  \end{figure}
  
In our data, tokens may distribute in different dialogue threads. 
Therefore, the relative distance between tokens cannot be calculated by subtracting their absolute position ids.
However, the RoPE uses the global index to represent relative distance:
\begin{equation}\nonumber
(\boldsymbol{\mathcal{R}}_m \boldsymbol{q})^{\top}(\boldsymbol{\mathcal{R}}_n \boldsymbol{k}) =  \boldsymbol{q}^{\top} \boldsymbol{\mathcal{R}}_m^{\top}\boldsymbol{\mathcal{R}}_n \boldsymbol{k} = \boldsymbol{q}^{\top} \boldsymbol{\mathcal{R}}_{n-m} \boldsymbol{k},
\end{equation}
where $m$ and $n$ are the absolute positions of two tokens.
And $m\text{-}n$ is not the relative distance of two tokens, which means the RoPE can not work typically.
Therefore, we develop a method to calculate the relative token distance for different thread pairs.
In detail, for each token $t$, we define the distance between $t$ and the root node as its local position id $P_t$.
For each two threads $T_i$ and $T_j$, the absolute positions to represent their relative distance can be calculated by the Algorithm~\ref{alg.rela}.

For example, as the block with different color shows in Fig.~\ref{fig:distance_matrix}, we can see that $(P^{ij}_t-P^{ij}_{t'})(t\in T_i, t'\in T_j)$ is the relative distance of $t$ and $t'$.
Then we use the calculated $P^{ij}_t$ as the absolute position to perform the RoPE operation.

\paragraph{Specification of Baselines.}

As no prior method is deliberately designed for DiaASQ, we consider re-implementing several strong-performing systems closely related to the task as our baselines.
Here we give a complete description on these baseline systems.

\begin{compactitem}
  \item \textbf{CRF-Extract-Classify} is a three-stage system (extract, filter, and combine) proposed for the sentence-level quadruple ABSA by~\citet{CaiXY20}. 
  Here we retrofit the model to further support \emph{target} term extraction.
  \item \textbf{SpERT} is proposed by~\citet{EbertsU20} for joint extraction of entity and relation based on a span-based transformer.
  Here we slightly modify the model to support triple-term extraction and polarity classification.
  \item \textbf{Span-ASTE} is a span-based approach for triplet ABSA extraction~\cite{XuCB20}.
  Similarly, we change it to be compatible with the DiaASQ task by editing the last stage of Span-ASTE to enumerate triplets.
  \item \textbf{ParaPhrase} is a generative seq-to-seq model for the quadruple ABSA extraction~\cite{ZhangD0YBL21}.
  We modify the model outputs to adapt to our DiaASQ task. 
\end{compactitem}

In particular, ParaPhrase~\cite{ZhangD0YBL21} is a generative model proposed for the quadruple ABSA task.
We re-implement the model and modify the output to fit it with our task.
In short, given the source dialogue, we expect the model to output a sentiment-aware string:

``\texttt{Target} is \textit{great/bad/ok}, because the \texttt{Aspect} of it is \texttt{Opinion} ...'', 

\noindent where \texttt{Target/Aspect/Opinion} is a term palace-holder, and \textit{greate/bad/ok} is an opinionated expression indicating the specific sentiment polarity, i.e.,  positive/negative/other.
For the dialogue in Fig.~\ref{fig:example}, a promising output is:

\textit{Xiaomi 6} is \textit{great} because the \textit{screen quality} of it is \textit{very nice}.

\begin{table}[!ht]
\fontsize{10}{11.5}\selectfont
  \centering
\begin{tabular}{cc} 
\hline
    Param. & Value\\
\hline
Learning rate(BERT) & 1e-5\\
Learning rate(Other) & 1e-3\\
Batch size & 4 (dialogues) \\
Max grad norm & 1.0 \\
Weight decay & 0.01 \\
Epoch size & 20 \\
$\theta$ & 10,000 \\
$\bm{\alpha}$  & [1, 5, 5, 5] \\
$\beta$ & 0.5 \\
$\eta$ & 0.5 \\
\hdashline
Parameter scale & 210M \\
Training time / epoch & 3min20s \\
\hdashline
CPU & Intel i9 \\
GPU & NVIDIA RTX 3090 \\
\hline  
\end{tabular}
\caption{Detail of the hyper-parameter setting.}
\label{table:hyper}
  \vspace{-10pt}
\end{table}

\paragraph{Hyper-Parameters.}
Here we detail the experimental setups.
The testing results are shown by our model tuned on the developing set to achieve the best developing performances.
Hyper-parameters are listed in Table~\ref{table:hyper}. 
We adopt AdamW as BERT optimizer. 
Our model is implemented with PyTorch and trained on the Ubuntu-20.04 OS with the Intel i9 CPU and NVIDIA RTX 3090 GPU. 

\section{Extended Data Specification}
\label{Extended Data Specification}

\begin{figure}[!h]
\centering

\subfloat[Chinese dataset]{\label{fig:polarity_suba}
\includegraphics[width=0.45\textwidth]{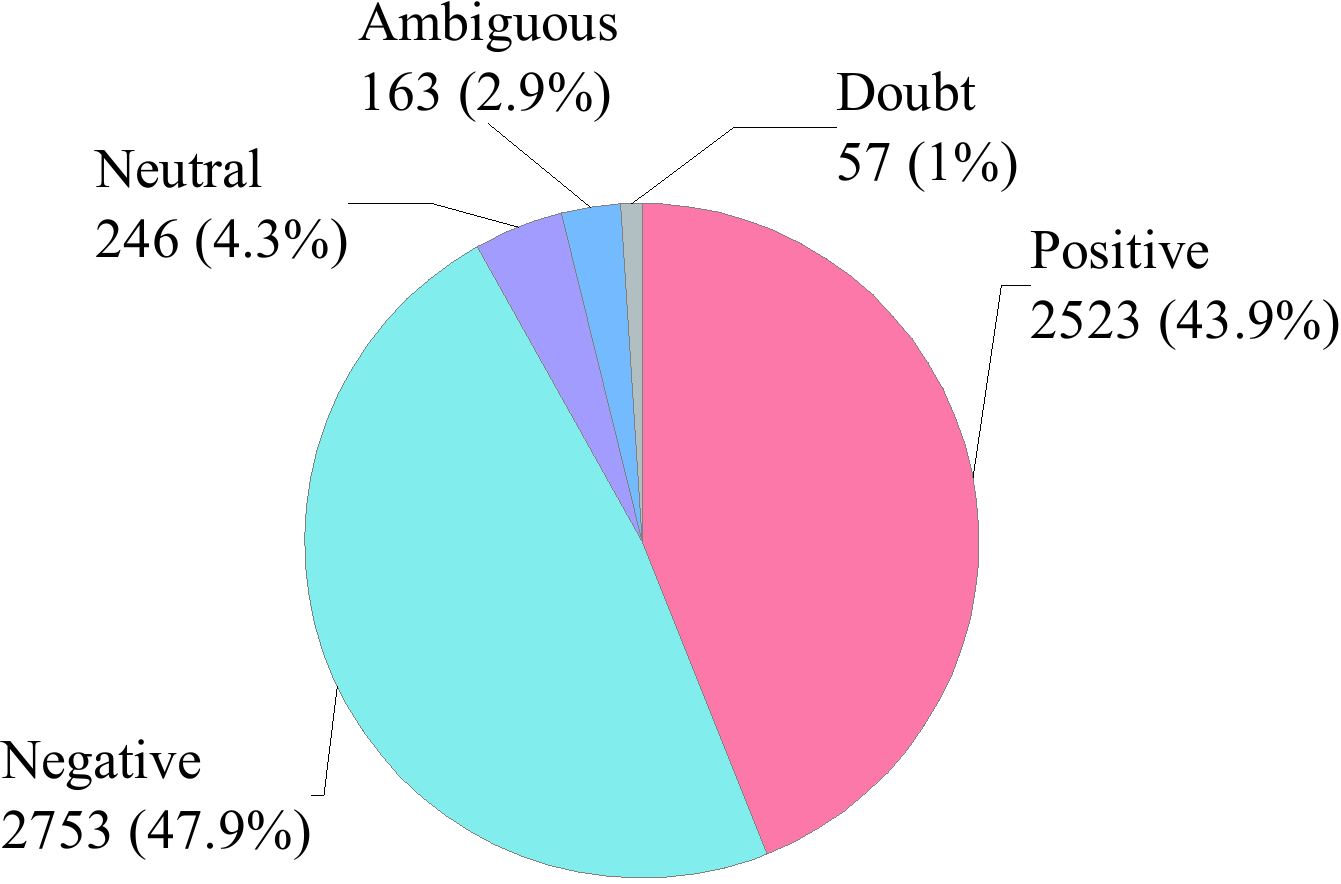}} \\  
	
\subfloat[English dataset]{\label{fig:polarity_subb}
\includegraphics[width=0.45\textwidth]{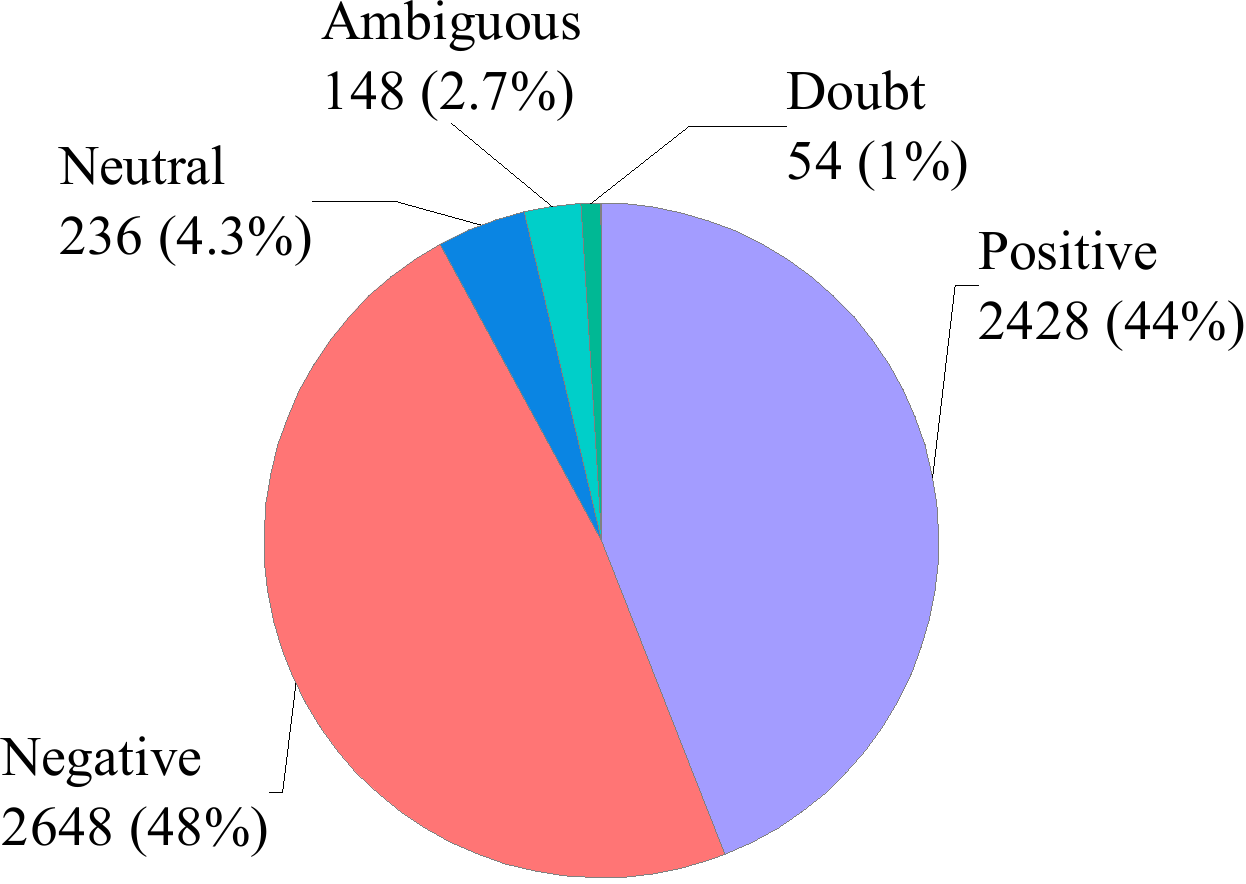}} \\ 

\caption{Distribution of quadruple polarities.}
\label{fig:polarity_distribution}
\end{figure}

\paragraph{Polarity Distribution.}
We statistics the polarity of quadruples in both Chinese and English datasets.
As illustrated in Fig.~\ref{fig:polarity_distribution}, most of the quadruple express the clear sentiment tendency, which is constituent with the users’ speaking habits on social media.
Then, the positive and negative sentiment rates are near, indicating that our data sampling is balanced.
Furthermore, due to the left three polarity being quite a few in our dataset, we merge them as a new category, others, for the convenience of extraction.

\begin{figure}[!t]
  \centering
  \includegraphics[width=0.46\textwidth]{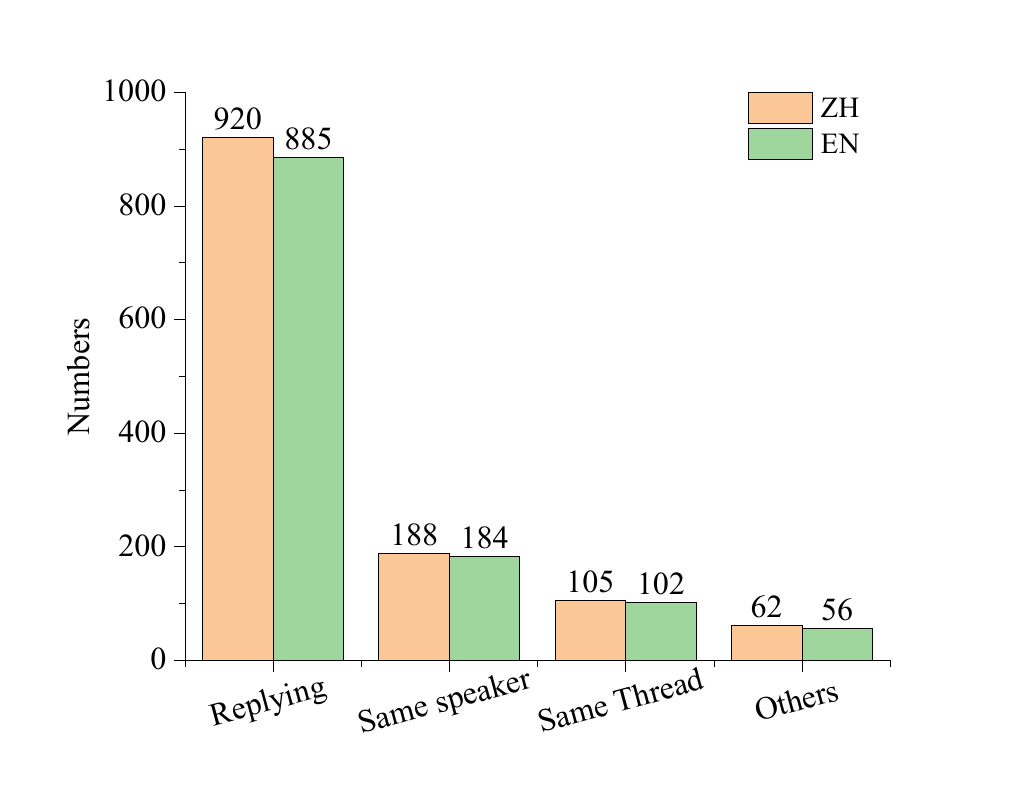}
  \caption{
  The number of different types of cross-utterance quadruples, whose elements at least come from two different utterances. 
  `Replying' denotes that the two utterances have replying relationship and `Same speaker' indicates the two utterances spoken by the same person.
  `Same thread' denotes the two utterances belonging to the same dialogue thread.
  `Others' mainly contains very rare cases, e.g., one quadruple contains elements from different threads.
  }
  \label{fig:cross-utterance}
\end{figure}

\paragraph{Cross-utterance Quadruples.}
We also analyzed the categories and numbers of cross-utterance quadruples.
As shown in Fig.~\ref{fig:cross-utterance}, most of the cross-utterance quadruples existed at two utterances next to each other, which reminds us that replying relationship can provide critical clues for cross-utterance quadruple extraction. 
Besides, the speaker and thread information needs to be further explored as they also indicate quite a few cross-utterance quadruples.

\begin{table}[!h]
\fontsize{9}{11.5}\selectfont
    \centering
\begin{tabular}{cccc} 
  \hline
      Lang & Index & Same Polarity & Different Polarity\\
  \hline
      \multirow{6}{*}{ZH} & 
      \myroman{1} & 90    & 71   \\
    & \myroman{2} & 1,684 & 1,366\\
    & \myroman{3} & 548   & 4    \\
    & \myroman{4} & 178   & 399  \\
    & \myroman{5} & 70    & 106  \\
    & \myroman{6} & 59    & 84   \\
  \hline  
      \multirow{6}{*}{EN} & 
      \myroman{1} & 102 & 67      \\
    & \myroman{2} & 1,578 & 1,260 \\
    & \myroman{3} & 549   & 4     \\
    & \myroman{4} & 170   & 382   \\
    & \myroman{5} & 67    &  99   \\
    & \myroman{6} & 62    & 77    \\
    \hline
  \end{tabular}
  \caption{Statistics of overlapped quadruples.
  The second column of each row is the index of the subplot in Fig.~\ref{fig:overlap}.
  }
  \label{fig:overlap-state}
  \end{table}

\begin{figure*}[!t]
  \centering
  \includegraphics[width=0.99\textwidth]{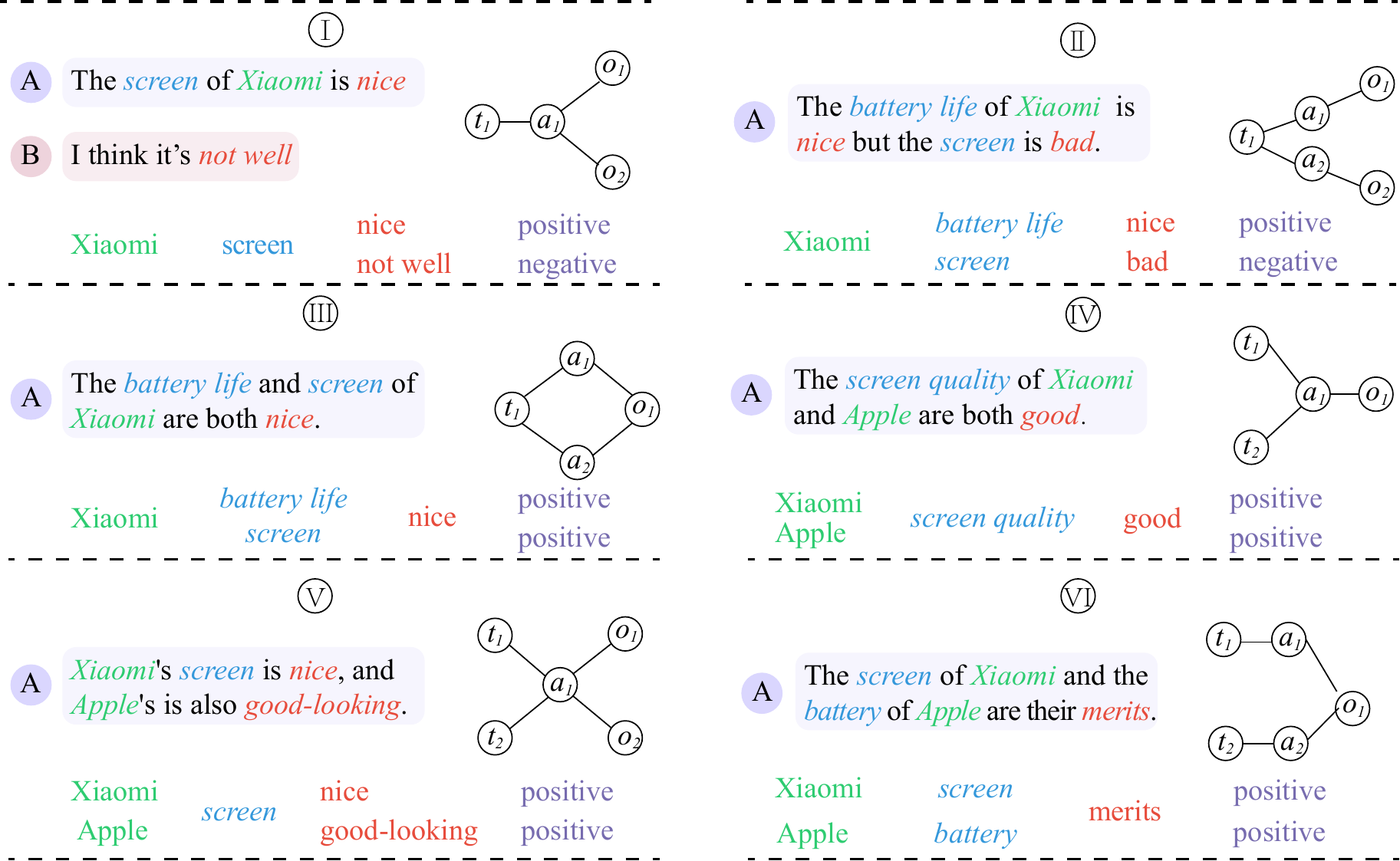}
  \caption{
  Quadruple overlap in our DiaASQ dataset, including a total of six cases.
  }
  \label{fig:overlap}
  \end{figure*}

\paragraph{Quadruple Overlapping.}
As we cast earlier, there are a good number of quadruples overlapping between each others in our DiaASQ dataset, which is not specially described in our main article due to space limitations.
As the first case shown in Fig.~\ref{fig:overlap}, two quadruples may contain the same target and opinion term.
The overlap information can actually provide valuable clues for better extraction.
Here we show in Fig.~\ref{fig:overlap} all types of overlap cases of the Chinese version dataset and their statistics information in Table~\ref{fig:overlap-state}.

\section{Specification on Data Construction}
\label{Specification on Data Acquisition}

This part describes the details that we constructed the DiaASQ dataset, including the data acquisition and annotation projection.

\begin{figure*}[!htbp]
  \centering
  \includegraphics[width=1.6\columnwidth]{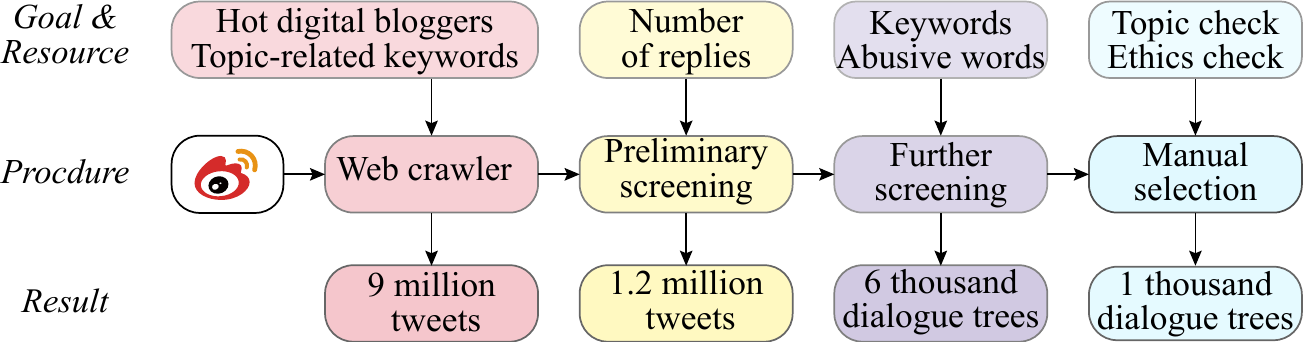}
  \caption{The workflow of data acquisition and preprocesssing.}
  \label{fig:crawler}
\end{figure*}

 \begin{figure*}[ht]
  \centering
  \includegraphics[width=1.85\columnwidth]{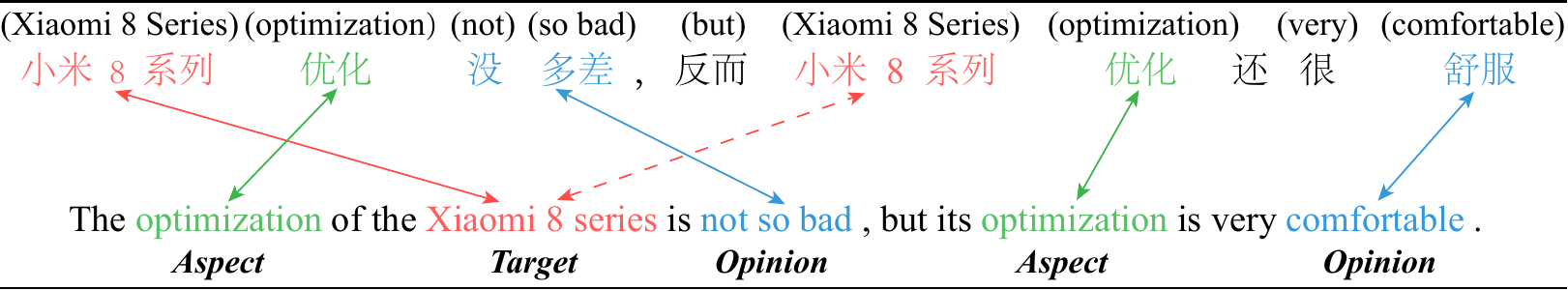}
  \caption{
  A example for projection correction.
  The red dotted line denote manually added alignment relation.
}
  \label{fig:alignment}
\end{figure*}
 
\subsection{Data Acquisition}

Fig.~\ref{fig:crawler} illustrates the overall workflow that we obtain a high-quality original corpus from social media.

First, based on the official leaderboard, we collected the top 100 influential digital-domain bloggers on Weibo and crawl their history tweets as many as possible.
Meanwhile, we also built a mobile-phone related keywords library and crawled tweets and their comment searched by these keywords.
After this step, we obtained nearly 9 million tweets and comments, and the replying relation was also recorded.
Then, we conduct a preliminary screening to exclude posts with less than ten replies or no father node.
About 1.2 million posts were retained after this procedure.
Next, according to the replies relation, we combine these posts into dialogue trees, whose root nodes are level-1 comments below each primary tweet, and the maximum depth is no more than 4.
Based on the phone-related keywords and a collection of abusive words, a more strict filtering rule is performed at the tree level.
In detail, a thread will be kept if it contains any two of the phone-related keywords and does not contain any abusive words.
Once a dialogue tree has three valid threads and the total number of nodes is between 6 and 10, it will be selected as the candidate dialogue. 
Around 6,000 dialogue trees are left after these steps.
Finally, we manually checked the candidate dialogue, and dialogues that are indeed phone-related and do have no ethical issue are selected as the final corpus.
After a very rigorous processing, we obtained 1,000 pieces of high-quality tree-like dialogues.

\begin{table}[!h]
\fontsize{9}{11.5}\selectfont
    \centering
\begin{tabular}{cc} 
  \hline
  Item & Text \\
  \hline
  Source&    \begin{CJK*}{UTF8}{gbsn}
\textit{所以我还是买了12x，虽然性价比不高}\end{CJK*} \\
  Translated &       \makecell[c]{\textit{So I still bought 12x,} \\ \textit{although the \textcolor{winered}{price} is not high}}\\
  Revision & \makecell[c]{\textit{So I still bought 12x,} \\ \textit{ although the \textcolor{mygreen}{cost-effective} is not high.}} \\
  \hline
  Source &  \begin{CJK*}{UTF8}{gbsn}
\textit{9带dc肯定香， 不过夏天你再试试} \end{CJK*} \\
  Translated &       \makecell[c]{\textit{9 with DC is definitely} \\ \textit{\textcolor{winered}{fragrant}, \textcolor{winered}{but you try } again in summer }}\\
  Revision &       \makecell[c]{\textit{9 with DC is definitely} \\ \textit{\textcolor{mygreen}{nice}, \textcolor{mygreen}{but you could try it } again in summer }}\\
  \hline  
  \end{tabular}
  \caption{
  Two typical translation revision examples.
  The first one is token-level translation error correction.
  And the second one shows a more proper statement.
  }
  \label{table:trainslate-error}
  \end{table}
 
\subsection{Parallel-Language Data Construction}
We also constructed an English version dataset based on the Chinese corpus via the annotation projection method.
Following~\citet{fei-etal-2020-cross} and~\citet{zhen-etal-2021-chinese}, the entire process contains two steps: text translation and annotation projection.
We manually revise the process result after each step to ensure the corpus quality.

\paragraph{Step1: Text Translation}
We first utilize Google Translate API to translate the Chinese text into English.\footnote{\url{https://cloud.google.com/translate}}
Despite the stunning performance of NMT(neural machine translation), it still makes some mistakes during translation.
The main reason is that our corpus is collected from social media and full of non-grammatical sentences, which has brought challenges for the NMT system to generate correct and elegant translations.
Therefore, we carefully revise the translation to eliminate errors and meanwhile improve readability.
Table~\ref{table:trainslate-error} lists one of the errors and revision results.

\paragraph{Step2: Annotation Projection}
Then we conduct projection to obtain English versions corpus based on original Chinese annotation.
Specifically, we achieve corpus projection with the help of awesome-align~\cite{dou-neubig-2021-word}, an excellent alignment tool based on large-scale multilingual language models.
We found that the alignment tool is not good at aligning named entities, and a representative error and correction are shown in Fig.~\ref{fig:alignment}.
After manually correcting all of the projection results, we obtained the final annotated corpus.

\subsection{Data Instances}
In Table~\ref{table:data-instance} we illustrate a full piece of data instance (a conversation) with our annotation (English version is shown).

\begin{table*}[!ht]
\renewcommand\arraystretch{1.35}
\fontsize{11}{11.5}\selectfont
    \centering
\begin{tabular}{l l p{7cm} p{6cm}} 
  \hline
  Key & \multicolumn{2}{l}{value}\\
  \hline
  Dialogue-ID & \multicolumn{2}{l}{0002}\\
  \hline
  \multirow{14}{*}{Dialogue}  & 0 & \multicolumn{2}{p{12.5cm}}{\textit{ This phone is not very good , but compared to the iPhone , I think it is better than the iPhone except for  the processor [ laughs cry ] } }\\
    & 1  &  \multicolumn{2}{p{12.5cm}}{\textit{ The iPhone is excellent as the processor and iOS , and others have been beaten by Android for many years .} }\\
    & 2 &  \multicolumn{2}{p{12.5cm}}{\textit{ Really . Sales also beat Android . Android manufacturers claim to be high - end and high - end every day , but they are just children in front of Apple .} }\\
    & 3 & \multicolumn{2}{p{12.5cm}}{\textit{  Samsung , Xiaomi does not all exceed Apple ? Because there are too many Android systems , there is only one iOS . If there is only one Android , what do you think of the result ? } }\\
    & 4 & \multicolumn{2}{p{12.5cm}}{\textit{  As you say , I have n't used Xiaomi , so I can 't comment . But traveling , my friend 's Xiaomi phone never took good photos . Especially when went to Malinghe Waterfall this week, we had to take pictures . Every photo taken by my brother 's Mi 11 was blurry . This experience is also speechless . } }\\
    & 5 & \multicolumn{2}{p{12.5cm}}{\textit{  Xiaomi 11 is really not good [ black line ] [ black line ] [ black line ] . } }\\
    & 6 & \multicolumn{2}{p{12.5cm}}{\textit{  The parameters overwhelm every year , and the experience is general every year ... that 's all . The phone is yours , who uses it , who knows . } }\\
  \hline
    Replies & \multicolumn{2}{l}{(-1, 0, 1, 2, 0, 4, 0, 6)}\\
    Speakers &  \multicolumn{2}{l}{(\ 0, 1, 2, 1, 3, 0, 3, 0)} \\
  \hline
      \multirow{4}{*}{Targets} & \multicolumn{2}{p{6cm}}{ (20, 21, \textit{iPhone})} & (30, 31, \textit{iPhone}) \\
    & \multicolumn{2}{p{6cm}}{ (82, 83, \textit{Samsung})} & (84, 85, \textit{Xiaomi}) \\
    & \multicolumn{2}{p{6cm}}{ (169, 171, \textit{Mi 11})} & (180, 182, \textit{Xiaomi 11}) \\
    & \multicolumn{2}{p{6cm}}{ (236, 237, \textit{iPhone})} & (248, 249, \textit{iPhone}) \\
  \hline
    \multirow{5}{*}{Aspects} & \multicolumn{2}{p{6cm}}{ (52, 53, \textit{Sales})} & (175, 176, \textit{experience}) \\
    & \multicolumn{2}{p{6cm}}{ (207, 208, \textit{experience})} & (199, 201, \textit{The parameters}) \\
    & \multicolumn{2}{p{6cm}}{ (35, 36, \textit{processor})} & (24, 25, \textit{processor}) \\
    & \multicolumn{2}{p{6cm}}{ (37, 38, \textit{iOS})} & (231, 232, \textit{experience}) \\
    & \multicolumn{2}{p{6cm}}{ (244, 246, \textit{image system})} & (145, 146, \textit{photos}) \\
  \hline
    \multirow{6}{*}{Opinions} & \multicolumn{2}{p{6cm}}{ (17, 18, \textit{better}, pos)} & (201, 202, \textit{ overwhelm}, pos) \\
    & \multicolumn{2}{p{6cm}}{ (54, 55, \textit{beat}, pos)} & (184, 186, \textit{not good}, neg) \\
    & \multicolumn{2}{p{6cm}}{ (172, 173, \textit{blurry}, neg)} & (32, 33, \textit{excellent}, pos) \\
    & \multicolumn{2}{p{6cm}}{ (178, 179, \textit{speechless}, neg)} & (88, 89, \textit{exceed}, pos) \\
    & \multicolumn{2}{p{6cm}}{ (209, 210, \textit{general}, neg)} & (243, 244, \textit{backward}, neg) \\
    & \multicolumn{2}{p{6cm}}{ (233, 234, \textit{better}, neg)} \\
  \hline
    \multirow{11}{*}{Quadruples} & \multicolumn{3}{l}{ (20, 21, 24, 25, 17, 18, pos, \textit{iPhone, processor, better})} \\
   &  \multicolumn{3}{l}{(30, 31, 35, 36, 32, 33, pos, \textit{iPhone, processor, excellent })} \\
   &  \multicolumn{3}{l}{(30, 31, 37, 38, 32, 33, pos, \textit{iPhone, iOS, excellent })} \\
   &  \multicolumn{3}{l}{(30, 31, 52, 53, 54, 55, pos, \textit{iPhone, Sales, beat })} \\
   &  \multicolumn{3}{l}{(82, 83, 52, 53, 88, 89, pos, \textit{Samsung, Sales, exceed })} \\
   &  \multicolumn{3}{l}{(84, 85, 52, 53, 88, 89, pos, \textit{Xiaomi, Sales, exceed })} \\
   &  \multicolumn{3}{l}{(180, 182, 145, 146, 184, 186, neg, \textit{Xiaomi 11, photos, not good })} \\
   &  \multicolumn{3}{l}{(248, 249, 244, 246, 243, 244, neg, \textit{iPhone, image system, backward })} \\
   &  \multicolumn{3}{l}{(236, 237, 231, 232, 233, 234, neg, \textit{iPhone, experience, better })} \\
   &  \multicolumn{3}{l}{(169, 171, 145, 146, 172, 173, neg, \textit{Mi 11, photos, blurry })} \\
   &  \multicolumn{3}{l}{(169, 171, 175, 176, 178, 179, neg, \textit{Mi 11, experience, speechless })} \\
  \hline  
  \end{tabular}
  \caption{
  An instance of our annotated corpus.
  The start and end positions of each entity are their global positions in the tokenized dialogue.
  `-1' in ``Replies” row indicate the corresponding utterance is the root of dialogue tree.
  }
  \label{table:data-instance}
  \end{table*}
\end{document}